%% file: main.tex
\documentclass[10pt,twocolumn,letterpaper]{article}

\usepackage[pagenumbers]{conf} 


\definecolor{confblue}{rgb}{0.21,0.49,0.74}
\usepackage[pagebackref,breaklinks,colorlinks,allcolors=confblue]{hyperref}

\def\confPaperID{550} 
\def\confName{conf}
\def\confYear{2027}

\title{CoRe-GS: Coarse-to-Refined Gaussian Splatting with Semantic Object Focus}

\author{
Hannah Schieber$^{1,2,3,4}$ 
Dominik Frischmann$^{1,2,3,4}$ 
Victor Schaack$^{1,2,3,4}$
Simon Boche$^{1,6}$ \and 
Angela P. Schoellig$^{1,4,5}$
Stefan Leutenegger$^{6}$
Daniel Roth$^{1,2,3}$ \\ 
$^{1}$Technical University of Munich, $^{2}$Human-Centered Computing and Extended Reality Lab\\
$^{3}$TUM University Hospital, Clinic for Orthopedics and Sports Orthopedics\\
$^{4}$Munich Institute of Robotics and Machine Intelligence (MIRMI)\\
$^{5}$Learning Systems and Robotics Lab (LSY), $^{6}$Mobile Robotics Lab (MRL), ETH Zürich\\
{\tt\small
\{firstname.lastname, vic.schaack\}@tum.de, stefan.leutenegger@mrl.ethz.ch}
}

\usepackage{xspace}
\def\approach{CoRe-GS\xspace}
\usepackage[nolist]{acronym}

\begin{document}
\input{acronym}
\maketitle
\begin{abstract}
Fast and efficient photorealistic 3D reconstruction with (semantic) Gaussian Splatting (GS) is crucial for time-critical robotic perception and navigation, where robots may need to rapidly reconstruct and inspect specific points of interest (POIs), such as  in search-and-rescue scenarios. Existing semantic GS methods prioritize reconstruction and segmentation quality while uniformly optimizing entire scenes, often overlooking runtime constraints crucial for real-world deployment. Moreover, object extraction often suffers from floaters and incomplete reconstructions, while post-optimization approaches  often produce incomplete POIs. We propose CoRe-GS, a coarse-to-refined, task-driven approach that first creates a segmentation-ready GS representation through lightweight late-stage semantic refinement, then selectively refines only the views and Gaussians associated with the POI. A color-based filtering strategy further suppresses segmentation-induced floaters. On LERF-Mask, our representation achieves competitive mIoU with 6-8$\times$ fewer optimization steps. On NeRDS 360 and SCRREAM, CoRe-GS reduces time-to-POI by 91-95\% over full-scene semantic GS while improving reconstruction quality and reducing floaters. Across all benchmarks, it is 44-76\% faster than post-optimization object extraction, demonstrating that task-aware refinement can deliver efficient, high-quality POI reconstruction for time-critical robotic perception and operation.
\end{abstract}

\section{Introduction}

\begin{figure}[t!]
    \centering
    \includegraphics[width=\columnwidth]{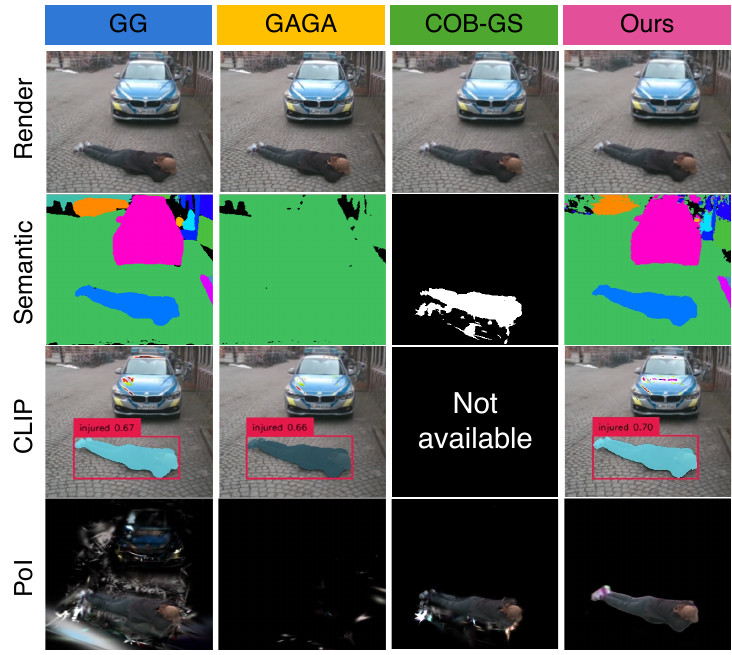}
    \caption{Our approach more consistently extracts the injured (staged) person as a \ac{poi} while requiring substantially less runtime (823s \ac{gg} \cite{ye_gaussian_2024}, 1558s GAGA \cite{lyu_gaga_2024}, \underline{348}s COB-GS \cite{zhang2025cob}, and \textbf{103}s ours). Notably, mask association fails for GAGA \cite{lyu_gaga_2024}, whereas our approach successfully extracts the \ac{poi}$^{1}$.}
    \label{fig:teaser}
\end{figure}

Capturing and reconstructing real-world environments in 3D is increasingly important for applications ranging from aerial imaging and disaster response \cite{li_cross-view_2025} to medicine \cite{yu_magnoramas_2021,kleinbeck_multi-layer_2024}, cultural heritage \cite{hasselman_arephotography_2023,schieber_semantics_controlled_2024}, and robotic
navigation \cite{tang_dronesplat_2025,schieber2026liftnav}. Representations such as
\ac{gs}~\cite{kerbl_3d_2023} provide detailed, explorable scene representation that support simulation, remote inspection, and robotic perception and planning \cite{chen2025splat,chen2025grad}.

In time-critical settings, however, reconstructing the complete scene at uniform quality is often unnecessary. Rescue teams or remote operators may require an accurate reconstruction of only one operationally relevant region, such as an injured person (see \autoref{fig:teaser}\footnote{Staged data captured with an Intel RealSense D456 RGB-D camera mounted on a custom drone platform. License plate anonymized in post-processing.}), a damaged structure, or an object to be manipulated. We refer to such a task-dependent semantic region as a \ac{poi}. A \ac{poi} is selected by the operator after coarse scene inspection, and its reconstruction quality directly affects the downstream task. Our objective is therefore not to obtain a fully converged semantic representation of the complete scene, but to minimize the time required to produce a high-fidelity reconstruction of the selected \ac{poi} (see \autoref{fig:teaser}).

Existing semantic \ac{gs} methods provide the scene understanding required to identify such regions, but commonly optimize semantic features throughout the complete scene optimization~\cite{ye_gaussian_2024,lyu_gaga_2024,schieber_semantics_controlled_2024,guo2026semantic,zhou2024feature}. This is inefficient and time-consuming when only a small subset of the scene remains relevant after operator selection. 

Active-view methods~\cite{li_activesplat_2025,jiang_fisherrf_2023} reduce acquisition cost by prioritizing informative viewpoints, but generally retain a scene-wide reconstruction objective. Object extraction and editing methods can isolate individual regions after reconstruction~\cite{ye_gaussian_2024,schieber_semantics_controlled_2024,hu_sagd_2024}, but residual inconsistencies between semantic masks and the Gaussian representation can leave background-associated Gaussians near the extracted object \cite{zhang2025cob}. During subsequent local optimization, these residuals may be amplified and appear as 3D outliers or floaters.

To address this setting, we introduce \approach, a task-conditioned coarse-to-fine \ac{gs} optimization strategy. \approach first constructs a lightweight, globally queryable scene representation, introducing semantic supervision only during a short late-stage phase. Once this representation supports semantic \ac{poi} selection, the operator selects a \ac{poi}, after which optimization is restricted to its associated views and Gaussians. Semantics therefore serve primarily to direct the remaining reconstruction budget toward the selected \ac{poi}, rather than to optimize the entire scene. To suppress floaters from retained background Gaussians, we introduce a scene-adaptive color-based filtering mechanism that identifies and periodically prunes background-associated Gaussians during local refinement, without semantic rasterization or multi-view mask back-projection.

We evaluate \approach in indoor and outdoor environments. We use LERF-Mask~\cite{ye_gaussian_2024,kerr_lerf_2023} to assess whether the lightweight initialization supports reliable semantic \ac{poi} selection. We then evaluate \ac{poi} reconstruction quality and optimization time on Tanks and Temples~\cite{knapitsch_tanks_2017}, SCRREAM~\cite{jung_scrream_2025}, and NeRDS 360~\cite{irshad_neo_2023}. Our experiments examine the trade-off among scene-wide semantic initialization cost, time-to-\ac{poi}, and final \ac{poi} reconstruction quality.

In summary, we contribute\footnote{Link to GitHub after acceptance}:
\begin{itemize}
\item A lightweight, globally queryable \ac{gs} representation that introduces semantic supervision only during a short late-stage optimization phase.
\item A task-conditioned refinement strategy that reallocates the remaining optimization budget to the views and Gaussians associated with an operator-selected \ac{poi} focusing on efficient training.
\item A scene-adaptive furthest-color filtering strategy that periodically identifies and removes segmentation-induced background Gaussians during local \ac{poi} refinement.
\end{itemize}


\section{Related Work}

Our work lies at the intersection of efficient scene reconstruction, semantic and editable Gaussian representations, and application-driven object-level reconstruction.

\subsection{Semantic Editing in Radiance Field Representations}

Several semantic and editable radiance-field approaches augment the scene representation with object labels \cite{schieber_semantics_controlled_2024}, learned feature fields\cite{ye_gaussian_2024,lyu_gaga_2024,qiu2024feature}, or language embeddings \cite{yu_language-embedded_2024,qin_langsplat_2024,zhang_3ditscene_2024}. Ye et al.~\cite{ye_gaussian_2024} introduce \ac{gg}. \ac{gg} learns a feature field that supports segmentation and editing, while GAGA~\cite{lyu_gaga_2024} employs a 3D-aware memory bank to improve grouping consistency. Aside from the usage of a memory bank, several approaches employ a contrastive learning scheme (e.g., \cite{silva_contrastive_2024,tang2026scar}) to improve label consistency of 2D segmentation foundation models \cite{kirillov_segment_2023,cheng_tracking_2023}. Schieber et al.~\cite{schieber_semantics_controlled_2024} direct label assignment instead of classifying a feature space, focusing primarily on the accurate editing of large-scale objects. More recently, several works focus on lifting features or semantics instead of learning them during rasterization  \cite{chacko2025lifting,marrie2025ludvig,hu_sagd_2024,guo2026semantic}. 

These representations enable downstream tasks such as semantic
search, and scene editing. Their optimization objectives commonly preserve semantic information across the complete scene so that multiple regions can be queried after reconstruction.

\subsection{Object-Centric Extraction and Refinement}

Early work on object-level manipulation focused on \ac{nerf}-based representations. For example, OR-NeRF~\cite{yin_or-nerf_2023} combines multi-view segmentation with image inpainting, while Weder et al.~\cite{weder_removing_2023} reconstruct the background after removing scene content. Similar strategies have since been extended to \ac{gs}. Huang et al.~\cite{huang_3d_2025} combine SAM-based segmentation with depth-consistent inpainting. Several other \ac{gs} approaches using semantics~\cite{ye_gaussian_2024,lyu_gaga_2024,schieber_semantics_controlled_2024} enable object manipulation through mask- or feature-guided Gaussian selection. Other approaches operate directly on the explicit \ac{gs} representation. Qu et al.~\cite{qu_drag_2025} introduce point-based scene editing, while physics-aware methods enable object interaction with applications in \ac{vr}~\cite{jiang2024vr,wang2025decoupledgaussian,schieber_splatxtract_2026}. Hu et al.~\cite{hu_sagd_2024} perform post-processing on existing RGB-only \ac{gs} representations by lifting 2D segmentation labels from \ac{sam}-DEVA~\cite{kirillov_segment_2023,cheng_tracking_2023} into 3D.

More recent work targets \ac{poi} extraction and local refinement in \ac{gs} representations. COB-GS~\cite{zhang2025cob} improves object-level segmentation boundaries through boundary-adaptive Gaussian splitting guided by SAM2-derived masks~\cite{ravi_sam_2024}. SplatXtract~\cite{schieber_splatxtract_2026} uses SAM2~\cite{ravi_sam_2024} for open-world \ac{poi} extraction and refinement, combining region queries with on-the-fly reconstruction from unposed images~\cite{meuleman2025fly}, but remains constrained in camera-motion freedom.

\subsection{Applications of Semantic Gaussian Splatting}

Semantic \ac{gs} support downstream tasks that require the localization, extraction, and manipulation of individual objects. For example, for manipulation, object-aware Gaussian representations~\cite{li_object-aware_2024,ji_graspsplats_2024} are used to identify task-relevant objects and infer grasp configurations.

Beyond robotics, SCGS~\cite{schieber_semantics_controlled_2024} enables class-specific object extraction and editing in large outdoor scenes using supervised segmentation models. Similary,  SplatXtract~\cite{schieber_splatxtract_2026} addresses a VR use case that extends this direction to open-world region-of-interest extraction and refinement using SAM2-based queries. In virtual environments, VR-GS~\cite{jiang2024vr} and DecoupledGaussian~\cite{wang2025decoupledgaussian} support explicit object separation and physics-based interaction adding mesh-based structures to the radiance fields.

\subsection{Research Gap}

Existing semantic \ac{gs} methods optimize scene-wide semantic or feature representations for downstream queries~\cite{lyu_gaga_2024,ye_gaussian_2024,schieber_semantics_controlled_2024}. Object-centric methods enable local refinement under different assumptions: SplatXtract~\cite{schieber_splatxtract_2026} reconstructs on-the-fly from unposed, motion-constrained streams, precluding an upfront queryable scene, while COB-GS~\cite{zhang2025cob} improves segmentation boundaries without focusing on optimization effort.
\approach instead first provides a lightweight, queryable scene from posed images and then focuses on refinement for a task-relevant \ac{poi}. Its core contribution is budget reallocation and efficient filtering, restricting optimization to the selected \ac{poi}'s views and Gaussians. A scene-adaptive furthest-color filter further suppresses background Gaussians adding floaters.

\begin{figure*}[t!]
    \centering
    \includegraphics[width=\textwidth]{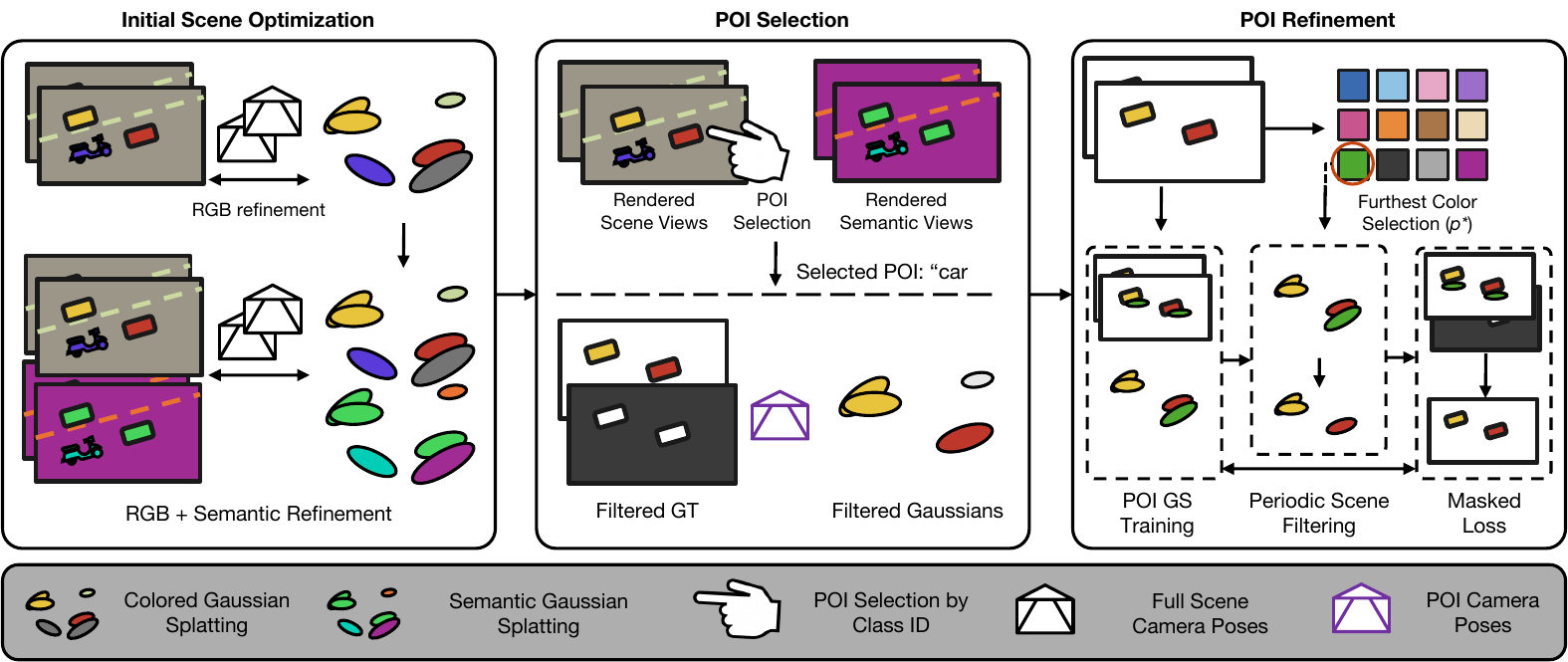}
    \caption{\textbf{\approach}. First, we produce a coarse \ac{gs} representation including a lightweight semantic optimization step (left). The \ac{poi} is extracted based on the semantic segmentation associations created in the previous step (center). Lastly, \ac{poi} refinement leverages our furthest color extraction ($p*$) prior training and periodic scene filtering using $p*$ during training (right).}
    \label{fig:approach}
\end{figure*}

\section{Method}

\approach combines fast \ac{gs} initialization~\cite{fang2024mini} with early-stage segmentation-aware refinement to create a semantic \ac{gs} representation. Building on this representation, operators can select semantic-dependent \acp{poi}, enabling interactive and local refinement and isolation of relevant points (see~\autoref{fig:teaser} and~\autoref{fig:approach}). Our refinement leverages color-based filtering to extract meaningful Gaussians and prevent outliers (floaters) enabling a floater-free \ac{poi} isolation.

\subsection{Initial Scene Optimization}
\label{sec:poi}

First, we purely optimize RGB \ac{gs} to recover coarse scene geometry from captured images. We leverage rapid densification, to achieve a fast and reliable reconstruction \cite{fang2024mini}. In parallel, semantic segmentation masks can be extracted (e.g., SAM-DEVA \cite{cheng_tracking_2023}) and associated with the input views.

Let $N_{\mathrm{img}}$ denote the number of training images and
$T$ the total number of optimization iterations. Our late-stage semantic refinement is performed during the final $4 \times N_{\mathrm{img}}$, i.e., for iterations $t \in [T - 4N, T]$. During that stage the Gaussian are augmented with object-level feature channels randomly initialized and fine-tuned using Cross-Entropy-Loss  jointly with the \ac{nvs} loss objective (\acs{ssim} and L1). We employ a $1\times1$ convolutional projection that maps the per-Gaussian object feature space to the semantic label space \cite{ye_gaussian_2024}. The number of output channels corresponds to the number of semantic classes obtained during the segmentation step.

\subsection{Point of Interest Selection}

The late-stage semantic refinement enables class-consistent Gaussian labels while preserving the efficiency and convergence behavior of the original \ac{gs} optimization. During pre-extraction, \acp{poi} are selected by class ID. Images containing the chosen class are retained, and only Gaussians predominantly associated with the selected \ac{poi} are preserved. For each retained image, a binary mask ($M$) indicates \ac{poi} and non-\ac{poi} regions. These binary masks ($M$) are used during the later refinement stage.

\subsection{Point-of-Interest Refinement}

Our \ac{poi} refinement is two-stage: we first compute the furthest per-scene color, then use it to filter Gaussians without segmentation rasterization or back projection, preserving the \ac{poi} while removing floaters.

\subsubsection{Furthest Color Extraction}

To mitigate unwanted outliers (floaters), we introduce a color-based filtering step that analyzes the color distribution of the input views. 
For computational efficiency, each image is downscaled by a factor of $0.5$ prior to processing.

Let the set of existing image colors be defined as
\begin{equation}
    C = \{c_i\}_{i=1}^{N_{\mathrm{color}}}, \quad c_i \in [0,1]^3 \subset \textbf{R}^3,
\end{equation}
where $N$ denotes the number of unique RGB color vectors in the downscaled image.

Furthermore, we define a reduced RGB color space
\begin{equation}
    P = \{p_j\}_{j=1}^{J}, \quad p_j \in [0,1]^3 \subset \textbf{R}^3,
\end{equation}
where $J = |P|$ denotes the number of discretely sampled color candidates. The set $P$ is constructed by uniformly sampling the RGB spectrum along each channel. We sample 26 colors per channel reducing the search space from $256^3$ to $26^3$, a value we determined empirically.

To efficiently evaluate distances between candidate colors and image colors, a KD-tree is constructed over $C$. We then determine the color $p^* \in P$ that maximizes the minimum Euclidean distance to all existing image colors:
\begin{equation}
    p^* = \arg\max_{p_j \in P} \left( \min_{c_i \in C} \| p_j - c_i \|_2 \right),
\end{equation}
where:
\begin{align}
     \| p_j - c_i \|_2 &=
    \sqrt{(r_{p_j} - r_{c_i})^2 +
           (g_{p_j} - g_{c_i})^2 +
           (b_{p_j} - b_{c_i})^2}
\end{align}

Here, $r$, $g$, and $b$ denote RGB intensities normalized to the interval $[0,1]$. The furthest color $p^*$ is used as a dedicated background color to identify and filter unwanted floaters during refinement.

\subsubsection{Periodic Scene Filtering}

In the pre-extraction step, we received the most distinct (``furthest'') color ($p^{*}$) relative to the colors present in the input views. In the subsequent refinement step, this color is used in the rendering step as background color and leveraged during filtering. We additionally define the average distance from all observed image colors to $p^{*}$:
\begin{equation}
    d_{\mathrm{avg}} = \frac{1}{N_{\mathrm{color}}} \sum_{c_i \in C} \|c_i - p^{*}\|_2,
\end{equation}
which serves as a scene-adaptive reference scale for the removal threshold used during periodic filtering.

\begin{figure*}[t!]
    \centering
    \includegraphics[width=\textwidth]{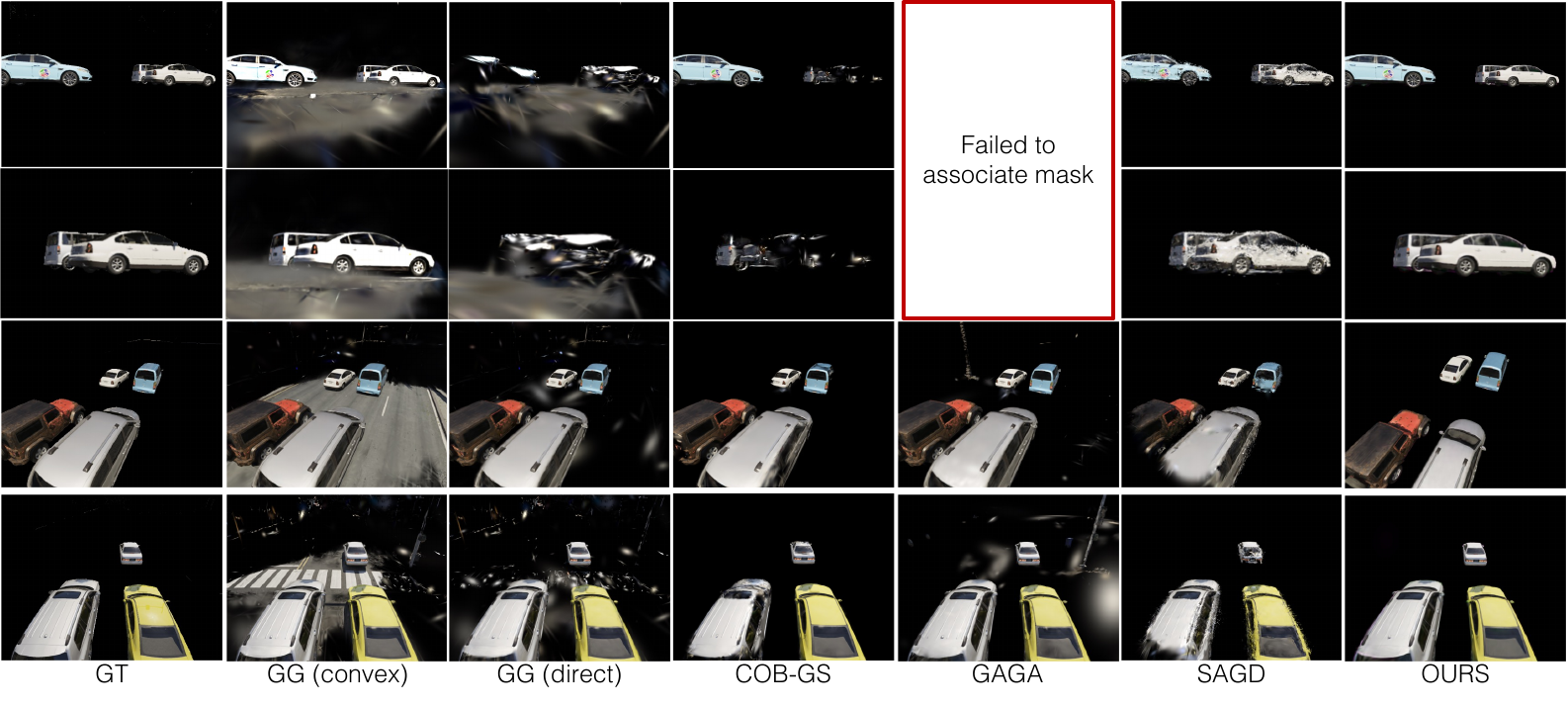}
    \caption{\textbf{\ac{poi} isolation.} We compare \ac{gg} \cite{ye_gaussian_2024}, COB-GS \cite{zhang2025cob}, GAGA \cite{lyu_gaga_2024}, SAGD \cite{hu_sagd_2024} and ours (\approach) on NeRDS360 \cite{irshad_neo_2023} (SF\_VanNessAveAndTurkSt5 (top), SF\_GrantAndCalifornia1 (bottom-center) and SF\_6thAndMission\_medium6 (bottom)). On SF\_VanNessAveAndTurkSt5 \ac{gg} shows several outliers, while COB-GS, shows less outliers but also a failure on the cars on the right side (see top row, and cropped image in top-center) GAGA fails during the mask association step for the memory bank.}
    \label{fig:nerds360}
\end{figure*}

Using $p^*$, we effectively filter out unwanted Gaussians as they get associated with the background color during training. Given the substantial color difference between $ p^*$ and meaningful Gaussians, outliers (floaters) can be identified and removed. Specifically, for each view, the color of each Gaussian is computed by evaluating its \ac{sh} coefficients relative to the camera's viewing direction. The resulting RGB color is then compared to the previously identified ($ p^*$). Then the Euclidean distance between the Gaussian's computed RGB color and  $p^*$ is calculated. If this distance falls below a threshold ($t_r$), the Gaussian is considered an artifact and marked for removal. Artifact Gaussians are pruned to suppress visual artifacts near object boundaries. Given the average distance ($d_{avg}$), the Gaussians ($G$) selected for removal $G_{\text{remove}}$ at a removal iteration are those satisfying:

\begin{equation}
    G_{\text{remove}} = \left\{ G_i \mid d(c_{G_i}, p^*) < \frac{d_{\text{avg}}}{2} \right\}.
\end{equation}

We apply this filtering every 1k iterations to suppress residual background artifacts.

\section{Evaluation}

We evaluate \ac{poi} quality using held-out views and ground-truth masks. We use black background for rendering in all approaches. We report \acs{psnr}, \acs{ssim}, and \acs{lpips}, mean masked \acs{psnr} and mean masked \acs{ssim}, total training time, and segmentation quality using \ac{miou} and \ac{mbiou}.

\subsection{Implementation Details}
\label{sec:impl}

$p^{*}$ is computed once in parallel using an $N$-worker thread pool. For initial training, we use 5k iterations, with the final $4\times$ the number of images iterations dedicated to segmentation rasterization. 
For \ac{poi} refinement, we use 18k iterations and apply color-based filtering every 1k iterations with a color-distance threshold multiplier of 0.5. All other parameters follow \cite{fang2024mini}. We use our binary masks ($M$) to mask the L1 loss. Further implementation details are provided in the supplementary material.
For controlled evaluation, we consider COLMAP-based \cite{schonberger2016structure} \ac{gs} approaches and exclude SLAM-based methods to ensure consistent and fair pose estimation and initialization.

\subsection{Datasets}

We evaluate our approach on four datasets. To assess visual quality we use Train and Truck  \cite{knapitsch_tanks_2017}. We obtained segmentation masks via SAM-DEVA \cite{cheng_tracking_2023}. Furthermore, we use the COLMAP scenes from NeRDS 360~\cite{irshad_neo_2023} with matching semantic ground truth. The scenes are particularly interesting as they show urban landscapes and vehicles, while offering a 360${^\circ}$ view of a street scene similar to drone captures. For an indoor proof of concept, we evaluate on scene \textit{1}, \textit{2}, and \textit{3} from SCRREAM \cite{jung_scrream_2025}. 

To assess segmentation quality, we use LERF-Mask~\cite{ye_gaussian_2024,kerr_lerf_2023} and follow Ye et al. \cite{ye_gaussian_2024} for the evaluation strategy on LERF-Mask.

\begin{table*}[t!]
         \caption{\textbf{Evaluation on NeRDS 360 \cite{irshad_neo_2023}.} We select ``car'' as \ac{poi}. ``-'' denotes a failed mask association. Best results are marked in \textbf{bold}. \label{tab:plain_NeRDS360}}
 \setlength{\tabcolsep}{1pt} 
    \resizebox{\textwidth}{!}{
    \begin{tabular}{l|ccc|ccc|ccc|ccc|ccc|ccc}
        \hline \hline
         & \multicolumn{3}{c|}{\textbf{\acs{gg}}~\cite{ye_gaussian_2024} (``direct'')} & \multicolumn{3}{c|}{\textbf{\acs{gg}}~\cite{ye_gaussian_2024} (convex)} & \multicolumn{3}{c|}{\textbf{COB-GS}~\cite{zhang2025cob} } & \multicolumn{3}{c|}{\textbf{GAGA}~\cite{lyu_gaga_2024} (direct)} & \multicolumn{3}{c|}{\textbf{SAGD}~\cite{hu_sagd_2024}} & \multicolumn{3}{c}{\textbf{Ours}} \\
         & {\acs{psnr}$\uparrow$} & {\acs{ssim}$\uparrow$} & {\acs{lpips}$\downarrow$} & {\acs{psnr}$\uparrow$} & {\acs{ssim}$\uparrow$}  & {\acs{lpips}$\downarrow$} & {\acs{psnr}$\uparrow$} & {\acs{ssim}$\uparrow$} & {\acs{lpips}$\downarrow$} & {\acs{psnr}$\uparrow$} & {\acs{ssim}$\uparrow$} & {\acs{lpips}$\downarrow$} & {\acs{psnr}$\uparrow$} & {\acs{ssim}$\uparrow$} & {\acs{lpips}$\downarrow$} & 
        {\acs{psnr}$\uparrow$} & {\acs{ssim}$\uparrow$} & {\acs{lpips}$\downarrow$}\\ 
        \hline 
        \multicolumn{3}{l}{6thAndMission\_medium} \\ \hline
        6  & 18.369 & 0.634 & 0.276 & 11.844 & 0.462 & 0.359 & 20.415 & 0.909 & 0.085 & 18.781 & 0.716 & 0.181  & 21.425 & 0.901 & 0.128  &\textbf{25.430} & \textbf{0.948} &	\textbf{0.080}  \\
        7  & 26.200 & 0.853 & 0.118  & 18.611 & 0.748 & 0.18 & 23.862 &	0.943 &	0.058 & {27.438} & {0.945} & {0.068}  & 25.422 &	0.936 &	0.076 & \textbf{29.057} &	\textbf{0.965}	& \textbf{0.053}
        \\
        10 & 19.140 & 0.737 & 0.229 & 8.372 & 0.345 & 0.416 & 19.652	& 0.899 &	0.095 & 24.084 & 0.868 & 0.145 & 20.227	& 0.888 &	0.136 & \textbf{29.727}	& \textbf{0.959} &	\textbf{0.074}
        \\ \hline
        \multicolumn{3}{l}{GrantAndCalifornia} \\ \hline
        1 & 23.185 & 0.757 & 0.194 & 10.703 & 0.451 & 0.336 & 21.926 & 0.912 &	0.089 & 20.750 & 0.694 & 0.143  & 22.453 &	0.891	&0.125 & \textbf{30.676} &	\textbf{0.962} &	\textbf{0.066}
         \\
        2 & 20.113 & 0.682 & 0.236 & 10.594 & 0.362 & 0.424 & 15.556 & 0.826 &	0.161  &  25.360 & 0.886 & 0.096  & 22.291	& 0.888 &	0.123 &\textbf{30.699} &	\textbf{0.957}	& \textbf{0.066} \\
        3 &
        23.200 & 0.778 & 0.178  & 10.393 & 0.407 & 0.361 & 19.011 &	0.901 &	0.097 & 22.673 & 0.735 & 0.165  & 22.707 &	0.907	& 0.105 &\textbf{29.606} &	\textbf{0.961}	 & \textbf{0.059}
        \\  \hline
        \multicolumn{3}{l}{VanNessAveAndTurkSt} \\ \hline
        3 & 14.130 &	0.750 &	0.233 &	14.130 &	0.641 &	0.293 & 22.686 & 	0.914 &	\textbf{0.090}	& -	& -	& -	 & 15.731 & 0.817 & 0.203 &\textbf{27.837}	 &\textbf{0.949}	& {0.097}  \\
        5 & 13.670	& 0.567 & 0.316 & 13.073 & 0.516 & 0.309 & 17.772 &	0.900 &	0.095 &	- & - &	-   & 22.927	& 0.916	& 0.124& \textbf{26.848}	& \textbf{0.961} &	\textbf{0.081}  \\  \hline 
        Mean & 19.751 & 0.720 & 0.222 & 12.215 & 0.491 & 0.335 & 20.110 & 	0.900	&0.096&  - & - &-   & 21.648 & 0.893 & 0.127 &  \textbf{28.713}	& \textbf{0.958} &	\textbf{0.072} \\  \hline \hline
    \end{tabular} }
\end{table*}

\begin{table}[t!]

\setlength{\tabcolsep}{1pt} 
\caption{\textbf{Runtime evaluation on all datasets.} Total time in
seconds from posed inputs to an isolated \ac{poi}, 
evaluation and preprocessing are excluded. Best results are in \textbf{bold}. 
\label{tab:runtime} }   
 \setlength{\tabcolsep}{1.2pt}
\resizebox{\columnwidth}{!}{
\begin{tabular}{l|ccccc}
\hline \hline

& \textbf{\ac{gg}}~\cite{ye_gaussian_2024}
& \textbf{COB-GS}~\cite{zhang2025cob}
& \textbf{GAGA}~\cite{lyu_gaga_2024}
& \textbf{SAGD}~\cite{hu_sagd_2024}
& \textbf{Ours} \\ \hline
& Time(s) & Time (s) & Time (s) & Time (s) & Time (s)  \\ \hline
Mean NeRDS 360
& 1802
& 477
& 2416
& 350
& \textbf{114} \\ \hline
Mean Train and Truck 
& OOM
& 901 
& -
& 674 
& \textbf{353} \\  \hline
Mean SCRREAM
& 2991
& 925
& -
& 478
& \textbf{267} \\\hline \hline

\end{tabular}
}
\end{table}

\begin{figure}[t!]
    \centering
    \includegraphics[width=\columnwidth]{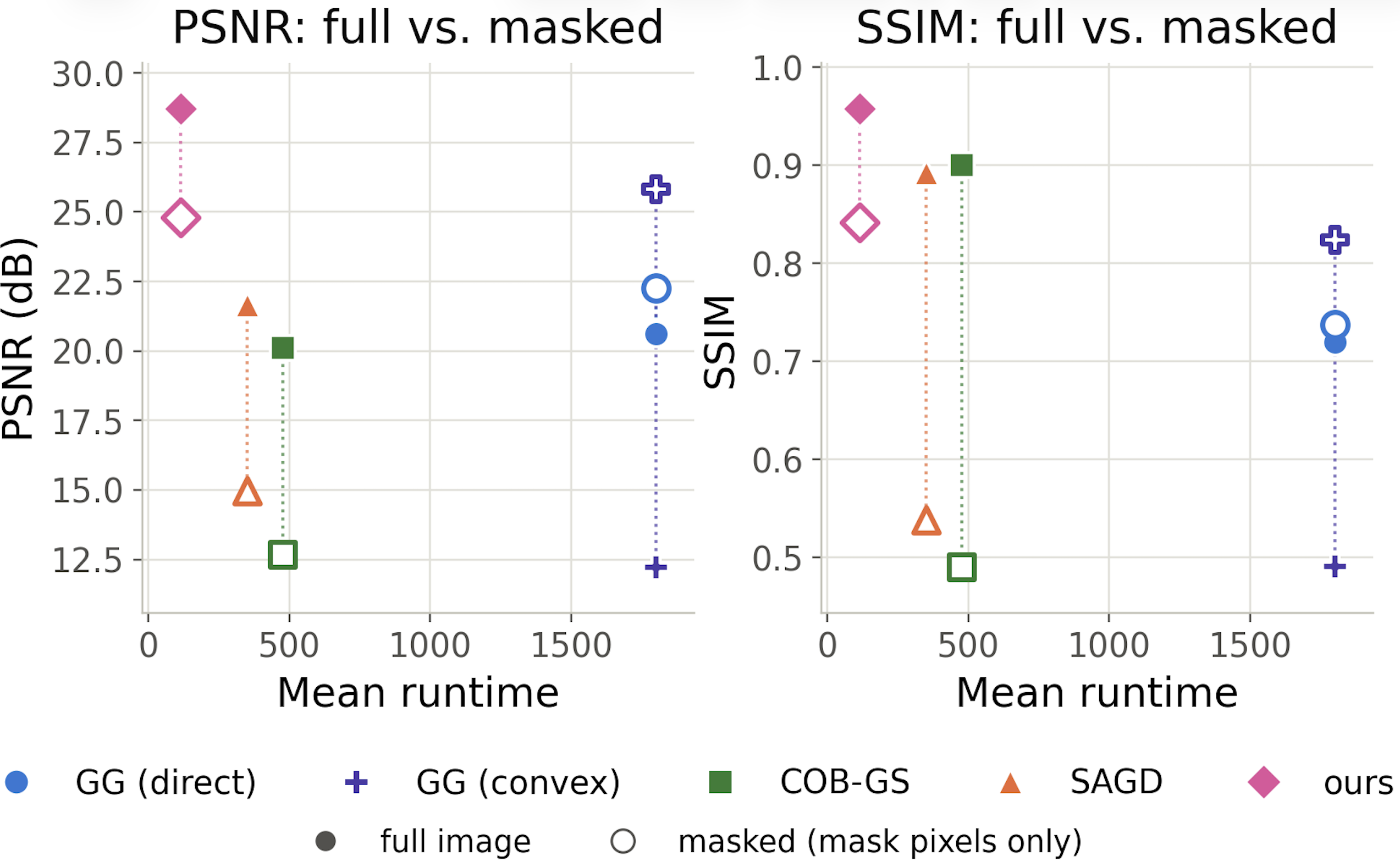}
    \caption{\textbf{Pareto on NeRDS 360.}  Runtime vs. (masked) metrics.}
    \label{fig:pareto}
\end{figure}

\begin{table*}[t!]

\caption{\textbf{Evaluation on Train and Truck~\cite{knapitsch_tanks_2017}.} We select ``train'' and ``truck'' as \ac{poi}.  Best results are \textbf{bold}.
\label{tab:tandt_psnr_ssim} }    

\centering
\setlength{\tabcolsep}{1pt} 
\resizebox{0.65\textwidth}{!}{
\begin{tabular}{l|ccc|ccc|ccc|ccc}
\hline \hline

& \multicolumn{3}{c|}{\textbf{\ac{gg} (direct)}~\cite{ye_gaussian_2024}}
& \multicolumn{3}{c|}{\textbf{COB-GS}~\cite{zhang2025cob}}
& \multicolumn{3}{c|}{\textbf{SAGD}~\cite{hu_sagd_2024}}
& \multicolumn{3}{c}{\textbf{Ours}} \\ \hline

& \acs{psnr}$\uparrow$ & \acs{ssim}$\uparrow$ & \acs{lpips}$\downarrow$
& \acs{psnr}$\uparrow$ & \acs{ssim}$\uparrow$ & \acs{lpips}$\downarrow$
& \acs{psnr}$\uparrow$ & \acs{ssim}$\uparrow$ & \acs{lpips}$\downarrow$
& \acs{psnr}$\uparrow$ & \acs{ssim}$\uparrow$ & \acs{lpips}$\downarrow$ \\ \hline

Train
& 17.462 & 0.662 & 0.231
& \textbf{19.144} & 0.703 & \textbf{0.185}
& 16.623 & 0.754 & 0.214
& 17.747 & \textbf{0.769} & 0.197 \\

Truck
& OOM
& OOM
& OOM
& 24.981 & 0.897 & 0.088
& 21.642 & 0.872 & 0.123
& \textbf{25.662} & \textbf{0.934} & \textbf{0.079} \\ \hline

Mean
& 17.462 & 0.662 & 0.231
& 22.063 & 0.800 & 0.137
& 19.133 & 0.813 & 0.169
& 21.705 & 0.852 & 0.138 \\ \hline \hline

\end{tabular}
}
\end{table*}

\subsection{Runtime}

Runtime follow the protocol of \autoref{sec:impl}: total wall-clock time from posed inputs to an isolated \ac{poi}, excluding rendering and the preprocessing shared by all methods. As denoted in \autoref{tab:runtime} our overall runtime is $\sim$114 seconds, while \ac{gg}, requires $\sim1802$ seconds in total and GAGA $\sim2416$ seconds (including the runs failed to associate mask) on NeRDS 360. GAGA was excluded for the other datasets due to failed mask association on NeRDS 360 (denoted in \autoref{tab:runtime} as -). As shown \autoref{fig:pareto} we outperform in the runtime and \ac{nvs} metric tradeoff. Adding segmentation to \ac{gs} can enhance quality while increasing training time. \approach only requires a segmentation-ready scene where around $\sim50$ seconds are attributed to the initial scene creation with segmentation labels and the remaining $\sim60$ seconds for the refinement and around 4 seconds to the \ac{poi} extraction on NeRDS 360. 

On ``Train'' and ``Truck'', \ac{gg} receives OOM. SAGD, COB-GS and ours, show successful results, see \autoref{tab:runtime}. On SCRREAM, our runtime is more influenced by the number of images including the \ac{poi} during refinement compared to the NeRDS 360 where most views contained the \ac{poi}. We outperform existing work \autoref{tab:runtime} and \autoref{tab:scrream_psnr_ssim}.

\begin{table*}[t!]    
\caption{\textbf{Evaluation on SCRREAM~\cite{jung_scrream_2025}.} We use the scenes 01/02/03 (\_full\_00) and select ``chair'' as \ac{poi}. Best results are \textbf{bold}.
\label{tab:scrream_psnr_ssim} }    

\centering
\setlength{\tabcolsep}{1pt} 
\resizebox{0.65\textwidth}{!}{    
\begin{tabular}{l|ccc|ccc|ccc|ccc}    
\hline \hline    
 & \multicolumn{3}{c|}{\textbf{\ac{gg} (direct)}~\cite{ye_gaussian_2024}} 
& \multicolumn{3}{c|}{\textbf{COB-GS}~\cite{zhang2025cob}} 
& \multicolumn{3}{c|}{\textbf{SAGD}~\cite{hu_sagd_2024}} 
& \multicolumn{3}{c}{\textbf{Ours}} \\ \hline    
~ & \acs{psnr}$\uparrow$  & \acs{ssim}$\uparrow$  & \acs{lpips}$\downarrow$  & \acs{psnr}$\uparrow$  & \acs{ssim}$\uparrow$  & \acs{lpips}$\downarrow$  & \acs{psnr}$\uparrow$  & \acs{ssim}$\uparrow$  & \acs{lpips}$\downarrow$  & \acs{psnr}$\uparrow$  & \acs{ssim}$\uparrow$  & \acs{lpips}$\downarrow$  \\ \hline    
01 & 45.521 & 0.973 & 0.018 & 30.561 & 0.934 &	0.046 & 34.184 & 0.972 & 0.031 &   \textbf{47.758} & \textbf{0.995} & \textbf{0.012} \\
02& 21.663 & 0.829 & 0.056  & 30.980 &  0.949 & 0.039 & 23.718 & 0.966 & 0.038  &   \textbf{ 36.397} & \textbf{0.984} & \textbf{0.015}  \\
03 & 46.929 & 0.964 & 0.018  & 26.649 & 0.992 & 	0.012 & 38.078 & 0.960 & 0.031 & \textbf{47.993} & \textbf{0.993} & \textbf{0.017}  \\ \hline
Mean & 38.038 & 0.922 & 0.031 & 29.397	& 0.958	& 0.032 & 31.994 & 0.966 & 0.033 & \textbf{44.043} & \textbf{0.991} & \textbf{0.015} \\ \hline \hline   
   
\end{tabular}    
}    
\end{table*}  

\subsection{Novel View Synthesis Quality}


\subsubsection{NeRDS 360}

As shown in \autoref{tab:plain_NeRDS360}, our approach outperforms \ac{gg}, GAGA, and SAGD. We outperform the baseline in all visual metrics. However, GAGAs mask association step fails on two scenes. In \autoref{fig:nerds360}, we provide a direct comparison of filtered outputs comparing the baselines and \approach. A comparison of \approach with \ac{gg} using ``convex hull'' removal reveals that the ``convex hull'' method retains numerous outliers (floater). Floater still remain when using ``direct'' removal in \ac{gg} or GAGA. Compared to post-processing only (SAGD \cite{hu_sagd_2024}), we retrieve clearer object boundaries and higher-quality \ac{poi} extracts. Since GAGA showed failed mask association, we continued evaluation with \ac{gg} and SAGD. For mean masked \acs{psnr}/\acs{ssim}, \ac{gg} retrieves with direct removal 22.244/0.738, with convex removal 25.817/0.824, COB-GS 12.659/0.490, SAGD 14.969/0.538 and ours 24.808/0.842, showing that our method not only mitigates outliers better, but remains high quality within the \ac{poi}. Although convex removal for \ac{gg} shows a higher masked PSNR, \autoref{fig:nerds360} and \autoref{tab:plain_NeRDS360} clearly show that outliers are still present, while \approach performs successfully visually and in both metrics. A full per-scene breakdown can be found in the supplementary.

\begin{figure}[t!]
    \centering
    \includegraphics[width=\columnwidth]{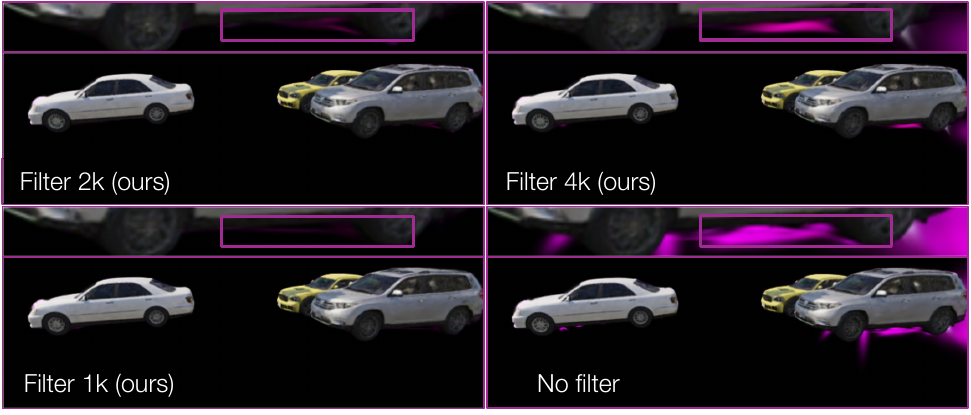}
    \caption{\textbf{Filtering.} No filtering shows outliers, filtering every 1k iter get the best result. Filtering every 2k/4k still produces outliers.}
    \label{fig:filtering}
\end{figure}

\subsubsection{Tanks and Temples}

On the "Truck" scene we outperform all baselines. On "Train", COB-GS achieves marginally higher PSNR and LPIPS while CoRe-GS achieves higher SSIM, see \autoref{tab:tandt_psnr_ssim}. For \ac{gg} we could not retrieve the results for the ``Truck'' scene on a RTX 4090 due to OOM. Still for SAGD, and \approach we could retrieve the result. We show a stronger performance in \acs{psnr}, compared to SAGD. Similarly for masked \acs{psnr}/\acs{ssim} we retrieve 23.052/0.833, while SAGD only achieves 15.546/0.617.

\subsubsection{SCRREAM}

As shown in \autoref{tab:scrream_psnr_ssim}, we outperform \ac{gg} and SAGD. Notably, in 02, we observe a substantial improvement of $+$14.7 dB in \acs{psnr} compared to \ac{gg}, indicating significantly enhanced robustness with fine-fidelity reconstructions of the chair as \ac{poi}. For mean masked \acs{psnr}/\acs{ssim} \ac{gg} retrieves with direct removal 33.200/0.878, SAGD 16.315/0.204 and ours 34.329/0.899, showing that our method not only mitigates outliers better and remains of high quality within \ac{poi}.

\subsection{Optimization Budget}

We investigate how the number of optimization iterations affects segmentation quality and whether a limited optimization budget is sufficient for reliable initialization. 

\subsubsection{Segmentation Quality}

To assess the segmentation quality after our initial optimization step, we compare \approach against state-of-the-art methods on the LERF-mask dataset using the evaluation strategy of \ac{gg}. As shown in \autoref{tab:lerf_mask}, \approach achieves the best performance on the teatime scene and remains competitive on figurines and ramen. On average, \approach achieves 73.8 \acs{miou} and 68.5 \acs{mbiou}, compared with 72.8 / 67.6 for \ac{gg}, 78.5 / 74.2 for GAGA, and 73.8 / 72.0 for COB-GS. Notably, \approach requires only 5k optimization iterations, compared to 30k for GG and 40k for GAGA and 30K +  $\delta$ for COB-GS.  $\delta$ denotes the number of finetuning iterations which vary in COB-GS \cite{zhang2025cob}. Thus, while we do not claim segmentation parity with fully converged scene-wide methods, our results demonstrate that a substantial fraction of their segmentation quality can be recovered at a small fraction of the optimization cost, which is sufficient for reliable POI selection (see \autoref{sec:init}).

\subsubsection{Initialization}
\label{sec:init}

Since \ac{poi} extraction and refinement depend on segmentation quality, we first assess initialization duration using LERF-Mask and Ye et al.~\cite{ye_gaussian_2024}. Increasing iterations from 5k to 8k leaves mean \acs{iou} unchanged at 73.8, while runtime increases from 77.7~s to 85.3~s, indicating no measurable segmentation benefit.

We further assess initialization duration for \ac{nvs} quality using \acs{psnr}, \acs{ssim}, and \acs{lpips} on \textit{SF\_6thAndMission\_medium6} from NeRDS 360~\cite{irshad_neo_2023}. Performance peaks at 5k iterations (25.43 \acs{psnr}, 0.948 \acs{ssim}, 0.080 \acs{lpips}) with a 17~s runtime, supporting 5k iterations as the initialization point.

\begin{figure}[t!]
    \includegraphics[width=\columnwidth]{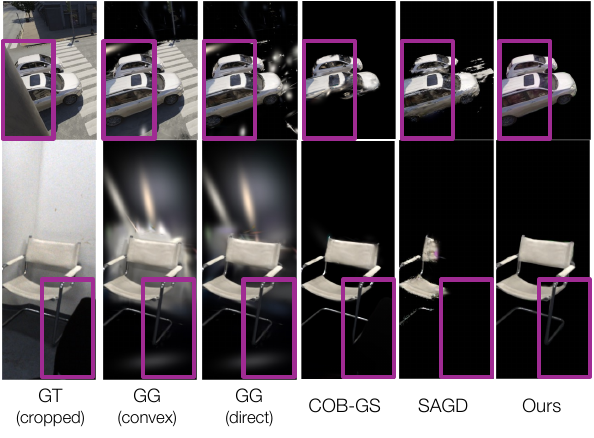}
    \caption{\textbf{Occlusion}. Light (bottom) and stronger (top) occlusion.}
    \label{fig:occlusion}
\end{figure}

\begin{table}[t!]

\caption{\textbf{Segmentation quality.} IoU on LERF-mask \cite{ye_gaussian_2024}. COB-GSs number of iterations varies depending on the number of images, denoted as $\delta$. \label{tab:lerf_mask}}

\setlength{\tabcolsep}{1pt}
\resizebox{\columnwidth}{!}{

\begin{tabular}{l|c|cccccc|cc}
\hline \hline
\textbf{Method} & iters &  \multicolumn{2}{c}{\textbf{figurines}}  & \multicolumn{2}{c}{\textbf{ramen}} & \multicolumn{2}{c}{\textbf{teatime}}& \multicolumn{2}{|c}{\textbf{Mean}}  \\
& & mIoU & mBIoU & mIoU & mBIoU & mIoU & mBIoU & mIoU & mBIoU \\
\hline 
LERF \cite{kerr_lerf_2023} & - & 33.5 & 30.6 & 28.3 & 14.7 & 49.7 & 42.6 & 37.2 &	29.3 \\
\ac{gg}~\cite{ye_gaussian_2024} & 30k & 69.7 & 67.9 & \underline{77.0} & \underline{68.7} & {71.7} & 66.1 & 72.8 & 67.6\\
GAGA \cite{lyu_gaga_2024} & 40k & \textbf{92.3} & \textbf{90.8} & 72.0 & 63.3 & 71.2 & {68.4} & \textbf{78.5} &	\textbf{74.2}\\
COB-GS \cite{zhang2025cob} & 30k + $\delta$ & 76.3 & 73.9 & \textbf{78.1} & \textbf{69.2} & \underline{77.2} & \underline{72.8} & \underline{77.2} & \underline{72.0}\\
Ours & 5k & \underline{80.1}	& \underline{76.7} &	63.8	& 54.2 &	\textbf{77.4} & \textbf{74.6} &  73.8 & 68.5  \\
\hline \hline

\end{tabular}
}

\end{table}

\subsection{Color-based Filtering}

In \autoref{fig:filtering}, we observe noticeable outliers without color-based filtering. Quantitatively, the results without filtering are 24.341 \acs{psnr}, 0.919 \acs{ssim}, and 0.093 \acs{lpips}. With our color-based filtering, \acs{psnr} improves to 25.430, \acs{ssim} improves to 0.948, and \acs{lpips} improves to 0.080. Aside from these quantitative metrics, we can visually observe clearer object boundaries in \autoref{fig:filtering}. 

\begin{figure}[t!]
    \begin{center}
        \includegraphics[width=\columnwidth]{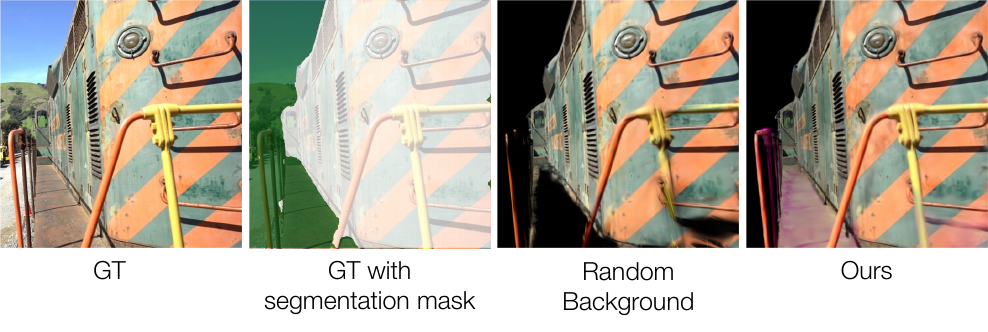}
    \caption{\textbf{Limitation.} Inaccurate GT segmentation masks (center, left) negatively affect \ac{poi} refinement for both random background and \approach. Still, our \approach (right) better preserves geometry, though color quality degrades.}
    \label{fig:limitation}
    \end{center}
\end{figure}

\paragraph{Random Background} Aside from our proposed work, random background colors are often used to omit floaters. However, as shown in \autoref{fig:limitation}, this leads to holes, while color-based filtering can lead to slight color-inconsistencies, but keeps the overall geometry more persistent.

\section{Discussion}

We propose \approach, a novel method for \ac{poi} refinement within \ac{gs}. Our hybrid strategy first constructs a segmentation-ready \ac{gs} representation at an early stage and subsequently performs high-quality refinements at selected \acp{poi}. This design substantially reduces overall training time (e.g., \autoref{tab:tandt_psnr_ssim}, \autoref{tab:plain_NeRDS360}, and \autoref{tab:scrream_psnr_ssim}). Importantly, \approach retains the ability to segment the complete scene. The segmentation performance is similar to existing approaches \cite{ye_gaussian_2024,lyu_gaga_2024}, as demonstrated in our evaluation (see Table \ref{tab:lerf_mask}). Furthermore, we demonstrate on the LERF-mask dataset, that relatively good segmentation performance can be achieved while relying on a tremendously lower number of optimization steps. For generating a segmentation-ready scene, our approach achieves a lower runtime of 50 and 136 seconds on NeRDS360 and SCRREAM respectively. Compared to pure post-processing approaches such as \cite{hu_sagd_2024}, our method produces visually more robust and coherent results (\autoref{fig:nerds360}). While alternative methods (e.g., \cite{wang2025decoupledgaussian}) enforce sharper boundaries via silhouette losses, they require a semantic rasterization step, an assumption incompatible with our runtime-efficient refinement stage. As demonstrated, the runtime of comparable semantic \ac{gs} approaches is notably higher. In contrast, our method provides a substantial speedup and extracts more visually appealing \acp{poi}. 

Our color-based filtering and \ac{poi} refinement enable an efficient and light-weight reconstruction, ensuring that only data from the target object contributes to the final scene. This results in a more focused and faster training by ignoring irrelevant background information. Challenges arise if segmentation masks are inaccurate. However, this challenge is general for approaches relying on the segmentation mask. Additionally \ac{poi} refinement gives \approach an advantage in occluded areas as shown in \autoref{fig:occlusion}. In the complete scene, a pole or street light can occlude the \ac{poi} from different \ac{nvs} perspectives. Semantic segmentation approaches \cite{ye_gaussian_2024,lyu_gaga_2024} leave floaters and post-processing only \cite{hu_sagd_2024} leads to a partly removed \ac{poi}.

Our main motivation lies in situations where operators can select \acp{poi} from emergency scenarios, which is why we focus mainly on outdoor data within the NeRDS 360 dataset. Experiments on Tanks and Temples are further added for demonstrating applicability on general \ac{gs} scenes. Furthermore, we demonstrate the applicability on self captured data (\autoref{fig:teaser}). Additional experiments on SCRREAM (indoor) validate the robustness of our method across diverse environments. 

\section{Limitations}

\approach benefits from high-quality segmentation. Its performance is therefore constrained by the segmentation quality of the underlying model, see  \autoref{fig:limitation}. This limitation is not unique to our approach. Previous work~\cite{schieber_semantics_controlled_2024} shows the impact of segmentation quality on \ac{nvs} quality and downstream tasks such as editing.

\section{Conclusion}

We introduced \approach, a \ac{poi} refinement strategy for \ac{gs} that selectively enhances segmentation-based \acp{poi}. By focusing computation on targeted scene areas, our approach produces higher-fidelity \acp{poi} without the overhead of refining the entire scene. We evaluate our segmentation-ready initialization and are in line with the state-of-the-art approaches. Furthermore, we evaluate \ac{nvs} quality on three benchmarks. We demonstrate clear advantages over state-of-the-art approaches. Our approach achieves targeted visual refinement and, compared multiple baselines, it reduces training time while preserving segmentation accuracy and rendering quality. These results highlight our approach’s efficiency and visual quality, while its modular design offers potential for future applications in more complex and specialized environments and mobile reconstruction scenarios.

\section*{Acknowledgment}

This work was partially supported by the Technical University of Munich, through its MIRMI Seed Fund program, and by the Robotics Institute Germany (RIG) under grant 16ME0997K. 
  
{
    \small
    \bibliographystyle{ieeenat_fullname}
    \bibliography{template}
}

\newpage
\section*{Appendix}
\section{Implementation Details}
\label{sec:impl}

$p^{*}$ is computed once in parallel using an $N$-worker thread pool. For initial training we use 5k iterations, with densification limited to the first 3k iterations, while aggressive cloning starts at iteration 500 and is performed every 250 iterations. We use a 3k-iteration warm-up, depth re-initialization at iteration 2k, and simplification at iteration 3k. The final $4\times$ the number of images iterations are dedicated to segmentation rasterization. 

For \ac{poi} refinement, we use 18k iterations with densification limited to 3k iterations, the same cloning schedule, depth re-initialization at 2k, simplification at 3k and 5k, and artifact filtering every 1k iterations with a color-distance threshold multiplier of 0.5. Other  parameters default to \cite{fang2024mini}. The filtering interval and threshold multiplier were selected via a parameter sweep (interval $\in$ {100, 500, 1k, 5k} iterations, threshold multiplier  $\in$  [0.5, 1.0]). Final PSNR/SSIM values were comparable across settings, while more frequent filtering slowed convergence and 1k filtering yielded the best results.

For controlled evaluation, we consider COLMAP-based \cite{schonberger2016structure} \ac{gs} approaches and exclude SLAM-based methods to ensure consistent pose estimation and initialization.

\section{Runtime}

We evaluate runtime across three datasets to assess the computational efficiency of our method against existing approaches. We first evaluate on NeRDS 360 \cite{irshad_neo_2023}, followed by Tanks and Temples \cite{knapitsch_tanks_2017} and SCRREAM \cite{jung_scrream_2025}, covering different scene types and capture conditions. For each dataset, we report the runtime required by each method, with lower values indicating faster execution.

\subsection{NeRDS 360}

For NeRDS 360, see \autoref{tab:runtime_comparison}, the runtime comparison shows that our method consistently achieves the lowest runtime across all evaluated scenes. With an average runtime of 114~s, our method is substantially faster than SAGD (349.5~s) \cite{hu_sagd_2024}, COB-GS (477~s) \cite{zhang2025cob}, \ac{gg} (1802~s) \cite{ye_gaussian_2024}, and GAGA (2416~s) \cite{lyu_gaga_2024}. This demonstrates a clear computational advantage while maintaining competitive \ac{nvs} quality.

\begin{table}[t!]
    \centering
    \caption{\textbf{Runtime comparison.} Runtime in seconds (Time $\downarrow$) for each method across the evaluated scenes. Lower is better.}
    \resizebox{\columnwidth}{!}{
    \begin{tabular}{l|ccccc}
        \hline \hline
        & \textbf{GG} \cite{ye_gaussian_2024} 
        & \textbf{COB-GS} \cite{zhang2025cob}
        & \textbf{GAGA} \cite{lyu_gaga_2024}
        & \textbf{SAGD}  \cite{hu_sagd_2024}
        & \textbf{Ours} \\

        & direct
        & refinement
        & direct
        & post-processing
        & refinement \\ \hline
        & Time $\downarrow$
        & Time $\downarrow$
        & Time $\downarrow$
        & Time $\downarrow$
        & Time $\downarrow$ \\ \hline
        \multicolumn{4}{l}{6thAndMission\_medium} \\ \hline
        6
        & 2055
        & 590
        & 1926
        & 473
        & \textbf{114} \\

        7
        & 1582
        & 429
        & 1348
        & 328
        & \textbf{107} \\

        10
        & 1662
        & 505
        & 3151
        & 350
        & \textbf{120} \\ \hline

        \multicolumn{4}{l}{GrantAndCalifornia} \\ \hline
        1
        & 1446
        & 459
        & 1409
        & 323
        & \textbf{126} \\

        2
        & 1579
        & 445
        & 1385
        & 351
        & \textbf{114} \\

        3
        & 2112
        & 429
        & 1377
        & 274
        & \textbf{101} \\ \hline
        \multicolumn{4}{l}{VanNessAveAndTurkSt} \\ \hline
        3
        & 2057
        & 488
        & 3890
        & 324
        & \textbf{114} \\

        5
        & 1920
        & 470
        & 4843
        & 373
        & \textbf{116} \\ \hline

        Mean
        & 1802
        & 477
        & 2416
        & 349.5
        & \textbf{114} \\ \hline \hline

    \end{tabular}
    }
    \label{tab:runtime_comparison}
\end{table}

\begin{table}[t!]
    \centering
    \caption{\textbf{Runtime comparison on Tanks and Temples.} Runtime in seconds (Time $\downarrow$) across the evaluated scenes. Lower is better.}
    \resizebox{\columnwidth}{!}{
    \begin{tabular}{l|cccc}
        \hline \hline
        & \textbf{GG} \cite{ye_gaussian_2024}
        & \textbf{COB-GS} \cite{zhang2025cob}
        & \textbf{SAGD} \cite{hu_sagd_2024}
        & \textbf{Ours} \\

        & direct
        & refinement
        & post-processing
        & refinement \\ \hline
        
        & Time $\downarrow$
        & Time $\downarrow$
        & Time $\downarrow$
        & Time $\downarrow$ \\ \hline

        Train
        & 2647
        & 845
        & 534
        & \textbf{413} \\

        Truck
        & 813
        & 958
        & 813
        & \textbf{293} \\ \hline

        Mean
        & 1730
        & 901
        & 673
        & \textbf{353} \\ \hline \hline

    \end{tabular}
    }
    \label{tab:runtime_comparison_tandt}
\end{table}

\begin{table}[t!]
    \centering
    \caption{\textbf{Runtime comparison.} Runtime in seconds across the evaluated scenes. Lower is better.}
    \resizebox{0.8\columnwidth}{!}{
    \begin{tabular}{l|cccc}
        \hline \hline
        & \textbf{GG} \cite{ye_gaussian_2024}
        & \textbf{COB-GS} \cite{zhang2025cob}
        & \textbf{SAGD} \cite{hu_sagd_2024}
        & \textbf{Ours} \\

        & Time $\downarrow$
        & Time $\downarrow$
        & Time $\downarrow$
        & Time $\downarrow$ \\ \hline

        Scene 1 & 2971 & 471 & 858 & \textbf{240} \\
        Scene 2 & 3027 & 496 & 1115 & \textbf{330} \\
        Scene 3 & 2976 & 468 & 802 & \textbf{203} \\ \hline

        Mean & 2991 & 478 & 925 & \textbf{267} \\ \hline \hline

    \end{tabular}
    }
    \label{tab:runtime_scrream}
\end{table}

\begin{table*}[t!]
    \centering
    \caption{\textbf{NVS quality comparison.} Comparison of masked PSNR (mPSNR) and masked SSIM (mSSIM). Higher values are better.}
    \resizebox{\textwidth}{!}{
    \begin{tabular}{l|cc|cc|cc|cc|cc|cc}
        \hline \hline
        & \multicolumn{2}{c|}{\textbf{\ac{gg}}~\cite{ye_gaussian_2024}}
        & \multicolumn{2}{c|}{\textbf{\ac{gg}}~\cite{ye_gaussian_2024}}       & \multicolumn{2}{c|}{\textbf{COB-GS}~\cite{zhang2025cob}}
        & \multicolumn{2}{c|}{\textbf{GAGA}~\cite{lyu_gaga_2024}}
        & \multicolumn{2}{c|}{\textbf{SAGD}~\cite{hu_sagd_2024}}
        & \multicolumn{2}{c}{\textbf{Ours}} \\
        
        & \multicolumn{2}{c|}{direct}
        & \multicolumn{2}{c|}{convex}
        & \multicolumn{2}{c|}{direct}
        & \multicolumn{2}{c|}{post-processing}
        & \multicolumn{2}{c}{} \\ \hline
        
        & mPSNR $\uparrow$ & mSSIM $\uparrow$ 
        & mPSNR $\uparrow$ & mSSIM $\uparrow$
        & mPSNR $\uparrow$ & mSSIM $\uparrow$
        & mPSNR $\uparrow$ & mSSIM $\uparrow$
        & mPSNR $\uparrow$ & mSSIM $\uparrow$
        & mPSNR $\uparrow$ & mSSIM $\uparrow$ \\ \hline
        
        6  & 23.905 & 0.821 & 26.987 & 0.852 &13.178 & 0.553 & 24.385 & 0.835 & 14.964 & 0.604 & \textbf{27.071} & \textbf{0.859} \\
        7  & 23.317 & 0.794 & \textbf{26.655} & \textbf{0.829} &15.623 & 0.630& 23.893 & 0.799 & 18.773 & 0.678 & 22.966 & 0.820 \\
        10 & 23.966 & 0.807 & \textbf{26.936} & 0.829 &11.695	& 0.429& 24.171 & 0.808 & 15.003 & 0.567 & 23.670 & \textbf{0.832} \\
        1  & 26.305 & 0.865 & \textbf{29.772} & \textbf{0.887} &15.462 & 0.644& 25.682 & 0.861 & 16.393 & 0.614 & 26.337 & 0.879 \\
        2  & 24.524 & 0.797 & \textbf{27.578} & 0.825 &8.114	& 0.151& 25.094 & 0.807 & 15.313 & 0.510 & 25.048 & \textbf{0.826} \\
        3  & 22.937 & 0.772 & \textbf{25.545} & \textbf{0.795} &10.472 & 0.363 & 22.769 & 0.769 & 14.412 & 0.443 & 22.393 & 0.790 \\
        3  & 24.465 & 0.831 & \textbf{27.091} & \textbf{0.859} &17.243 & 0.712& - & - & 9.254 & 0.309 & 24.859 & 0.855 \\
        5  & 8.533 & 0.214 & 15.968 & 0.713 &9.487	& 0.442& - & - & 15.640 & 0.577 & \textbf{26.121} & \textbf{0.874} \\ \hline
        
        Mean & 22.244 & 0.738 & \textbf{25.817} & 0.824 &12.659	& 0.490& 24.332 & 0.813 & 14.969 & 0.538 & 24.808 & \textbf{0.842} \\ \hline \hline
        
    \end{tabular}
    }
    \label{tab:nvs_comparison}
\end{table*}

\begin{table*}[t!]
    \centering
    \caption{\textbf{NVS quality comparison on Tanks and Temples.} Comparison of masked PSNR (mPSNR) and masked SSIM (mSSIM).}
    \begin{tabular}{l|cc|cc|cc|cc}
        \hline \hline
        & \multicolumn{2}{c|}{\textbf{\ac{gg}}~\cite{ye_gaussian_2024}}
        & \multicolumn{2}{c|}{\textbf{COB-GS}~\cite{zhang2025cob}}
        & \multicolumn{2}{c|}{\textbf{SAGD}~\cite{hu_sagd_2024}}
        & \multicolumn{2}{c}{\textbf{Ours}} \\
        
        & mPSNR $\uparrow$ & mSSIM $\uparrow$
        & mPSNR $\uparrow$ & mSSIM $\uparrow$
        & mPSNR $\uparrow$ & mSSIM $\uparrow$
        & mPSNR $\uparrow$ & mSSIM $\uparrow$ \\ \hline
        
        Train
        & 19.400 & 0.758 
        & 20.8100 &	0.8030
        & 17.600 & 0.723
        & \textbf{21.667} & \textbf{0.800} \\
        
        Truck
        & - & - 
        & 22.330 &	0.814
        & 13.493 & 0.511
        & \textbf{24.437} & \textbf{0.867} \\ \hline
        
        Mean
        & 19.400 & 0.758 
        & 21.570 &	0.809
        & 15.546 & 0.617
        & \textbf{23.052} & \textbf{0.833} \\ \hline \hline
        
    \end{tabular}
    \label{tab:nvs_tandt}
\end{table*}

\begin{table*}[t!]
    \centering
    \caption{\textbf{Masked NVS quality comparison on SCRREAM.} Comparison of masked PSNR (mPSNR) and masked SSIM (mSSIM).}
    \begin{tabular}{l|cc|cc|cc|cc}
        \hline \hline
        & \multicolumn{2}{c|}{\textbf{\ac{gg}}~\cite{ye_gaussian_2024}}
        & \multicolumn{2}{c|}{\textbf{COB-GS}~\cite{zhang2025cob}}
        & \multicolumn{2}{c|}{\textbf{SAGD} \cite{hu_sagd_2024}}
        & \multicolumn{2}{c}{\textbf{Ours}} \\

        & mPSNR $\uparrow$ & mSSIM $\uparrow$
        & mPSNR $\uparrow$ & mSSIM $\uparrow$
        & mPSNR $\uparrow$ & mSSIM $\uparrow$
        & mPSNR $\uparrow$ & mSSIM $\uparrow$ \\ \hline

        Scene01
        & 34.243 & 0.903 
        & 14.956 & 0.034 
        & 18.292 & 0.319
        & \textbf{35.140} & \textbf{0.906} \\

        Scene02
        & 27.248 & 0.794 & 19.427 & 0.716
        & 8.390 & 0.260
        & \textbf{29.707} & \textbf{0.796} \\

        Scene03
        & 38.109 & 0.939 & 30.438 & 0.752
        & 22.264 & 0.034
        & \textbf{38.140} & \textbf{0.993} \\ \hline

        Mean
        & 33.200 & 0.878 & 21.607 & 0.500
        & 16.315 & 0.204
        & \textbf{34.329} & \textbf{0.899} \\ \hline \hline

    \end{tabular}
    \label{tab:nvs_scrream}
\end{table*}

\subsection{Tanks and Temples}

On Tanks and Temples, see \autoref{tab:runtime_comparison_tandt}, our method consistently achieves the lowest runtime across both scenes, with a mean runtime of 353~s compared to 673~s for SAGD, 901~s for COB-GS, and 1730~s for \ac{gg}. This demonstrates a substantial computational advantage, reducing runtime by approximately 48\% compared to SAGD and 80\% compared to \ac{gg}.

\subsection{SCRREAM}

Our method achieves the lowest runtime across all three scenes, with a mean runtime of 267~s compared to 478~s for SAGD and 2991~s for \ac{gg}. This corresponds to a 44\% reduction compared to SAGD and a 91\% reduction compared to \ac{gg}.

\section{Masked Metrics}

To evaluate reconstruction quality independently of background regions and unwanted content, we additionally report masked NVS metrics. The evaluation focuses on the target \ac{poi}, allowing us to assess how accurately each method reconstructs the relevant scene content. We report masked PSNR and masked SSIM, with higher values indicating better reconstruction quality. 

\subsection{NeRDS 360}

On NeRDS 360, our method achieves competitive masked NVS quality, with a mean PSNR of 24.808 and SSIM of 0.842. While \ac{gg} with convex optimization achieves a higher mPSNR, our method achieves the highest mean SSIM among the evaluated approaches. More importantly, these masked metrics do not fully capture the advantage of our approach: on the non-masked metrics, our method performs substantially better, indicating improved overall scene reconstruction quality, see \autoref{fig:pareto}. This suggests that our approach not only reconstructs the target regions accurately but also effectively suppresses unwanted Gaussians and floaters outside the target region, resulting in a cleaner and more faithful overall reconstruction.

\subsection{Tanks and Temples}

\begin{figure}[t!]
    \centering
    \includegraphics[width=\columnwidth]{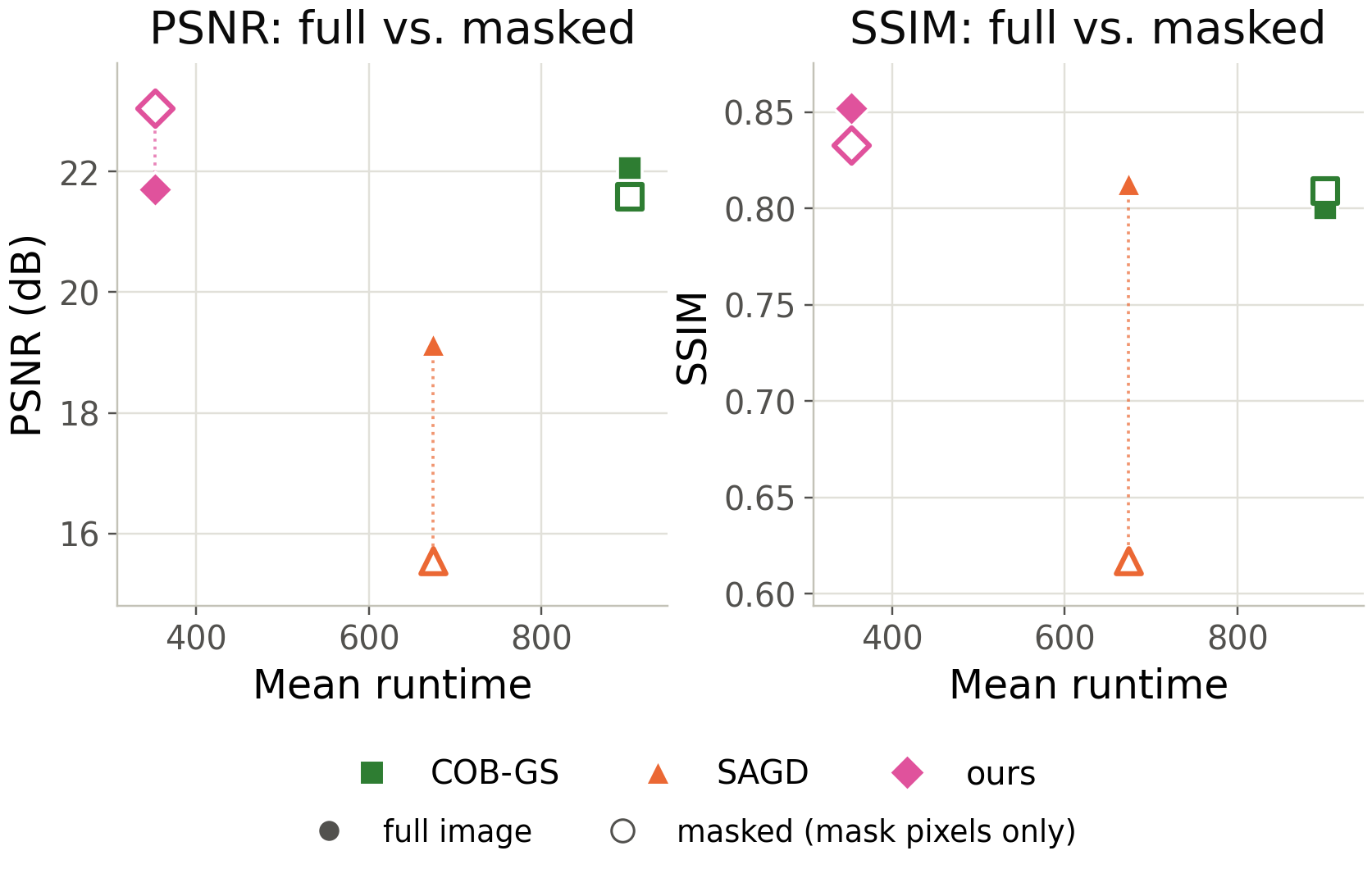}
    \caption{\textbf{Pareto on Tanks and Temples (Train and Truck).}  Runtime vs. (masked) PSNR and (masked) SSIM. While COB-GS shows overall good metrics, \approach stays consistent across masked and non-masked metrics with a lower runtime.}
    \label{fig:pareto_tt}
\end{figure}

On Tanks and Temples, see \autoref{tab:nvs_tandt}, our method achieves the best overall \ac{nvs} quality, with a mean mPSNR of 23.052 and SSIM of 0.833, substantially outperforming both \ac{gg} (19.400 / 0.758) and SAGD (15.546 / 0.617), see \autoref{fig:pareto_tt}. The improvement is particularly pronounced on the Truck scene, where our method reaches 24.437 PSNR and 0.867 SSIM, compared to 13.493 and 0.511 for SAGD. These results further indicate that our approach produces cleaner reconstructions with fewer unwanted Gaussians and floaters.

\subsection{SCRREAM}

\begin{figure}[t!]
    \centering
    \includegraphics[width=\columnwidth]{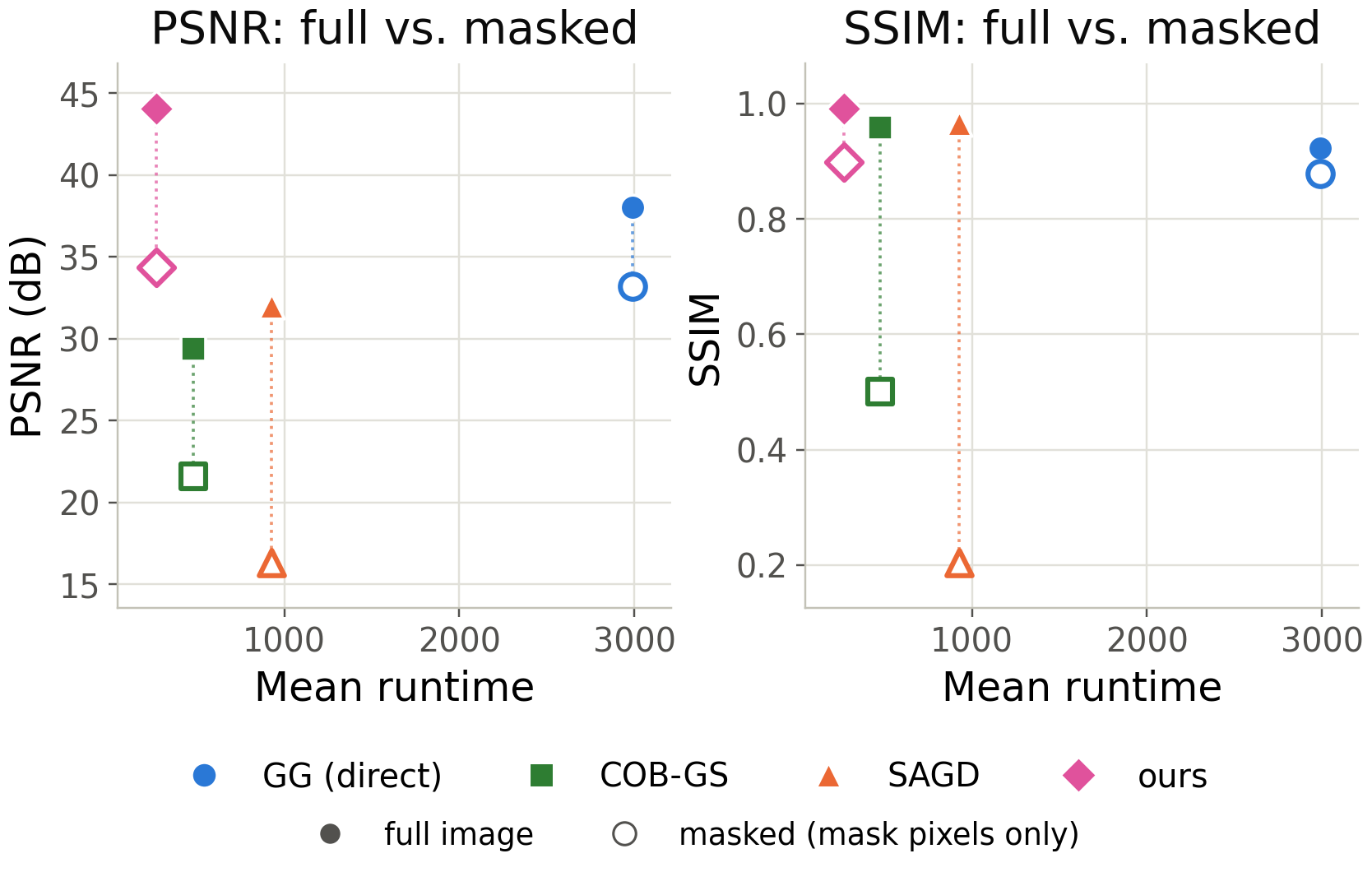}
    \caption{\textbf{Pareto on SCRREAM.}  Runtime vs. (masked) PSNR and (masked) SSIM. \approach shows clear advantages in both metrics.}
    \label{fig:pareto_scream}
\end{figure}

On SCRREAM, our method consistently achieves the highest masked \ac{nvs} quality across all three scenes, see \autoref{tab:nvs_scrream}. Furthermore, in the masked vs non-masked comparison, we outperform the other approaches, see \autoref{fig:pareto_scream}.

\section{Initialization}

Since both \ac{poi} extraction and later refinement is dependent on the segmentation quality we test segmentation quality on the LERF-Mask dataset, following the evaluation strategy from Ye et al. \cite{ye_gaussian_2024}. As shown in Table \ref{tab:init_iou}, the impact on \ac{iou} performance with a prolonge runtime is not measurable.

In addition to that, we measured the number of initial iterations using  \ac{psnr}  and  \ac{ssim} to cross validate the sufficiency of 5k iterations for initialization (see Table \ref{tab:init_nerds}). We run experiments on the scene \textit{SF\_6thAndMission\_medium6}.

\begin{figure}[t!]
    \centering
    \includegraphics[width=\columnwidth]{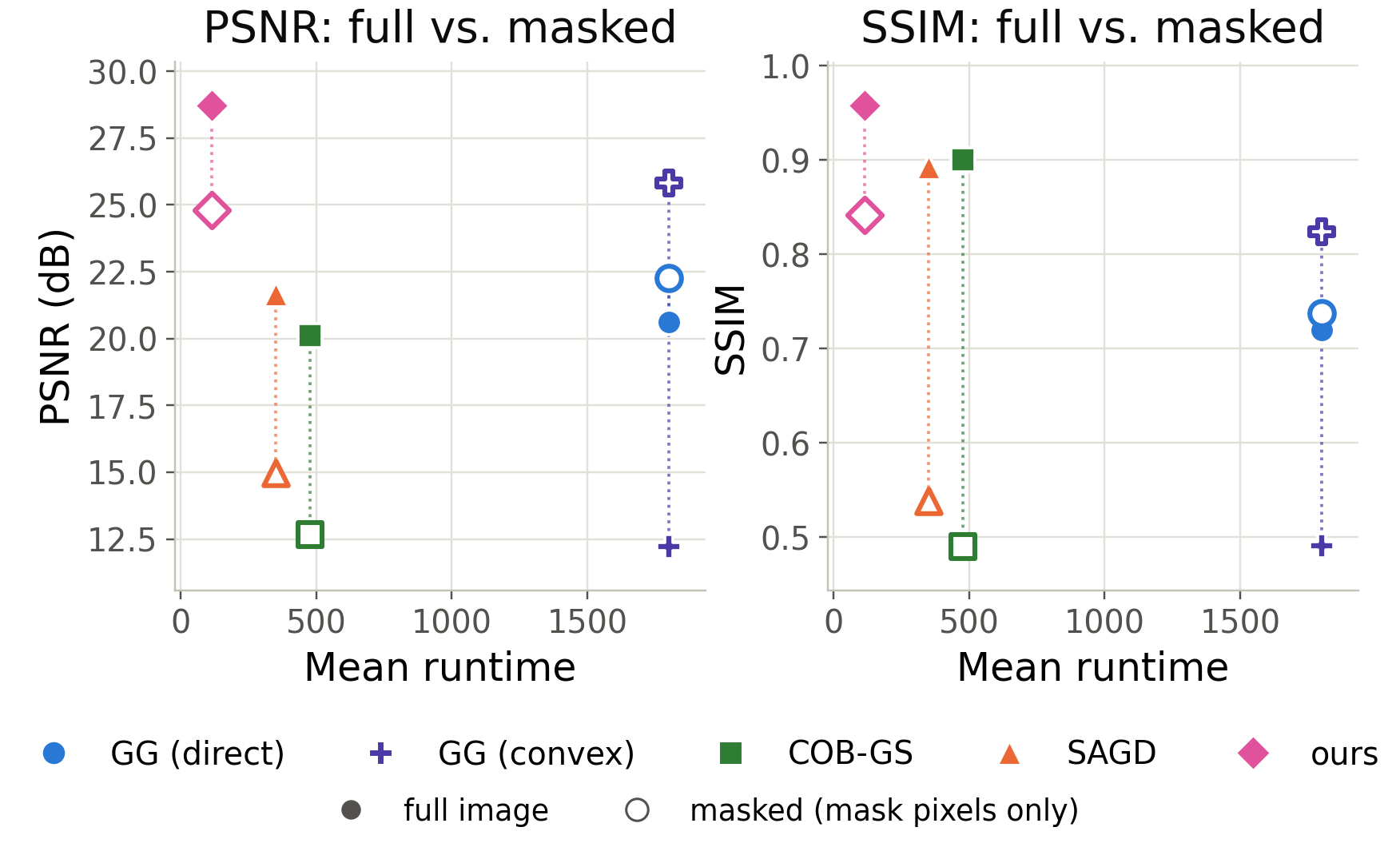}
    \caption{\textbf{Pareto on NeRDS 360.}  Runtime vs. (masked) PSNR and (masked) SSIM.}
    \label{fig:pareto}
\end{figure}

\begin{table}[t!]
    \caption{\textbf{IoU vs Time.} Comparison on LERF-mask~\cite{ye_gaussian_2024}.}
    \centering
    \resizebox{0.58\columnwidth}{!}{
    \begin{tabular}{l|cc} \hline \hline
       Iteration  & Mean \acs{iou} $\uparrow$ & TIME $\uparrow$ \\ \hline
       5000  & 73.8 & 77.7  \\
       6000  & 71.9 & 78.8 \\
       7000  & 73.0 & 80.3 \\
       8000  & 73.8 & 85.3 \\ \hline \hline
    \end{tabular}}
    \label{tab:init_iou}
\end{table}

\begin{table}[t!]
    \caption{\textbf{Initialization \ac{nvs}.} Comparison on NeRDS 360 \cite{irshad_neo_2023}. We compare \ac{nvs} quality vs runtime across different numbers of iterations.}
    \centering
     \resizebox{0.75\columnwidth}{!}{\begin{tabular}{l|ccccc} \hline \hline
Iters & TIME $\uparrow$		& PSNR  $\uparrow$ &	SSIM	$\uparrow$ & LPIPS $\uparrow$ \\  \hline
1000 &	 \textbf{9}	&23.619	&0.941	&0.096 \\ 
2000	&10	&23.144	&0.914	&0.101 \\
3000	&13	&25.055	&0.948	&0.080 \\
4000	&15	&25.011	&0.947	&0.081 \\
5000&	17	& \textbf{25.430}	& \textbf{0.948}&	\textbf{0.080} \\ \hline \hline
    \end{tabular}}
    \label{tab:init_nerds}
\end{table}

\section{Summary}

The supplementary material provides additional evidence for the efficiency, reconstruction quality, and robustness of \approach. Across NeRDS 360, Tanks and Temples, and SCRREAM, our method consistently achieves the lowest runtime, demonstrating substantial computational advantages over existing approaches. We further evaluate masked NVS quality to isolate reconstruction performance within the target \ac{poi}. Our method achieves competitive or superior masked PSNR and SSIM while showing stronger performance on non-masked metrics, indicating improved overall reconstruction quality and effective suppression of unwanted Gaussians and floaters.

Finally, we analyze the initialization process and show that 5,000 iterations provide sufficient segmentation and reconstruction quality. Increasing the initialization budget does not yield measurable improvements in \ac{iou}, while the NVS evaluation confirms that 5,000 iterations provide the best reconstruction quality among the tested configurations. Together, these results support the efficiency and effectiveness of our coarse-to-refined reconstruction strategy.

\end{document}

%% file: acronym.tex
\begin{acronym}[Bspwwww.]  

\acro{ar}[AR]{augmented reality}
\acro{ap}[AP]{average precision}
\acro{api}[API]{application programming interface}
\acroplural{ann}[ANN]{artifical neural networks}
\acro{bev}[BEV]{bird eye view}
\acro{rbob}[BRB]{Bottleneck residual block}
\acroplural{rbob}[BRBs]{Bottleneck residual blocks}
\acro{mbiou}[mBIoU]{mean Boundary Intersection over Union}
\acro{poi}[POI]{Point of Interest}
\acro{cai}[CAI]{computer-assisted intervention}
\acro{ce}[CE]{cross entropy}
\acro{cad}[CAD]{computer-aided design}
\acro{cnn}[CNN]{convolutional neural network}

\acro{crf}[CRF]{conditional random fields}
\acro{dpc}[DPC]{dense prediction cells}
\acro{dla}[DLA]{deep layer aggregation}
\acro{dnn}[DNN]{deep neural network}
\acroplural{dnn}[DNNs]{deep neural networks}

\acro{da}[DA]{domain adaption}
\acro{dr}[DR]{domain randomization}
\acro{fat}[FAT]{falling things}
\acro{fcn}[FCN]{fully convolutional network}
\acroplural{fcn}[FCNs]{fully convolutional networks}
\acro{fov}[FoV]{field of view}
\acro{fv}[FV]{front view}
\acro{fp}[FP]{False Positive}
\acro{fpn}[FPN]{feature Pyramid network}
\acro{fn}[FN]{False Negative}
\acro{fmss}[FMSS]{fast motion sickness scale}
\acro{gan}[GAN]{generative adversarial network}
\acroplural{gan}[GANs]{generative adversarial networks}
\acro{gcn}[GCN]{graph convolutional network}
\acroplural{gcn}[GCNs]{graph convolutional networks}
\acro{gs}[GS]{Gaussian Splatting}
\acro{gg}[GG]{Gaussian Grouping}
\acro{hmi}[HMI]{Human-Machine-Interaction}
\acro{hmd}[HMD]{Head Mounted Display}
\acroplural{hmd}[HMDs]{head mounted displays}
\acro{iou}[IoU]{intersection over union}
\acro{irb}[IRB]{inverted residual bock}
\acroplural{irb}[IRBs]{inverted residual blocks}
\acro{ipq}[IPQ]{igroup presence questionnaire}
\acro{knn}[KNN]{k-nearest-neighbor}
\acro{lidar}[LiDAR]{light detection and ranging}
\acro{lsfe}[LSFE]{large scale feature extractor}
\acro{llm}[LLM]{large language model}
\acro{map}[mAP]{mean average precision}
\acro{mc}[MC]{mismatch correction module}
\acro{mis}[MIS]{Minimally Invasive Surgery}
\acro{msdl}[MSDL]{Multi-Scale Dice Loss}
\acro{ml}[ML]{Machine Learning}
\acro{mlp}[MLP]{multilayer perception}
\acro{miou}[mIoU]{mean Intersection over Union}
\acro{nn}[NN]{neural network}
\acroplural{nn}[NNs]{neural networks}
\acro{ndd}[NDDS]{NVIDIA Deep Learning Data Synthesizer}
\acro{nocs}[NOCS]{Normalized Object Coordiante Space}
\acro{nerf}[NeRF]{Neural Radiance Fields}
\acro{NVISII}[NVISII]{NVIDIA Scene Imaging Interface}
\acro{ngp}[NGP]{Neural Graphics Primitives}
\acro{or}[OR]{Operating Room}
\acro{pbr}[PBR]{physically based rendering}
\acro{psnr}[PSNR]{peak signal-to-noise ratio}
\acro{pnp}[PnP]{Perspective-n-Point}
\acro{rv}[RV]{range view}
\acro{roi}[RoI]{region of interest}
\acroplural{roi}[RoIs]{region of interests}
\acro{rbab}[BB]{residual basic block}
\acro{ras}[RAS]{robot-assisted surgery}
\acroplural{rbab}[BBs]{residual basic blocks}
\acro{spp}[SPP]{spatial pyramid pooling}
\acro{sh}[SH]{spherical harmonics}
\acro{sgd}[SGD]{stochastic gradient descent}
\acro{sdf}[SDF]{signed distance field}
\acro{sfm}[SfM]{structure-from-motion}
\acro{sam}[SAM]{Segment-Anything}
\acro{sus}[SUS]{system usability scale}
\acro{ssim}[SSIM]{structural similarity index measure}
\acro{sfm}[SfM]{structure from motion}
\acro{slam}[SLAM]{simultaneous localization and mapping}
\acro{tp}[TP]{True Positive}
\acro{tn}[TN]{True Negative}
\acro{thor}[thor]{The House Of inteRactions}
\acro{tsdf}[TSDF]{truncated signed distance function}
\acro{vr}[VR]{Virtual Reality}
\acro{ycb}[YCB]{Yale-CMU-Berkeley}

\acro{ar}[AR]{augmented reality}
\acro{ate}[ATE]{absolute trajectory error}
\acro{bvip}[BVIP]{blind or visually impaired people}
\acro{cnn}[CNN]{convolutional neural network}
\acro{c2f}[c2f]{coarse-to-fine}
\acro{fov}[FoV]{field of view}
\acro{gan}[GAN]{generative adversarial network}
\acro{gcn}[GCN]{graph convolutional Network}
\acro{gnn}[GNN]{Graph Neural Network}
\acro{hmi}[HMI]{Human-Machine-Interaction}
\acro{hmd}[HMD]{head-mounted display}
\acro{mr}[MR]{mixed reality}
\acro{iot}[IoT]{internet of things}
\acro{llff}[LLFF]{Local Light Field Fusion}
\acro{bleff}[BLEFF]{Blender Forward Facing}

\acro{lpips}[LPIPS]{learned perceptual image patch similarity}
\acro{nerf}[NeRF]{neural radiance fields}
\acro{nvs}[NVS]{novel view synthesis}
\acro{mlp}[MLP]{multilayer perceptron}
\acro{mrs}[MRS]{Mixed Region Sampling}

\acro{or}[OR]{Operating Room}
\acro{pbr}[PBR]{physically based rendering}
\acro{psnr}[PSNR]{peak signal-to-noise ratio}
\acro{pnp}[PnP]{Perspective-n-Point}
%
\acro{sus}[SUS]{system usability scale}
\acro{ssim}[SSIM]{similarity index measure}
\acro{sfm}[SfM]{structure from motion}
\acro{slam}[SLAM]{simultaneous localization and mapping}

\acro{tp}[TP]{True Positive}
\acro{tn}[TN]{True Negative}
\acro{thor}[thor]{The House Of inteRactions}
\acro{ueq}[UEQ]{User Experience Questionnaire}
\acro{vr}[VR]{virtual reality}
\acro{who}[WHO]{World Health Organization}
\acro{xr}[XR]{extended reality}
\acro{ycb}[YCB]{Yale-CMU-Berkeley}
\acro{yolo}[YOLO]{you only look once}
\end{acronym}

%% file: main.bib
@String(CVPR= {IEEE Conf. Comput. Vis. Pattern Recog.})

@String(ICCV= {Int. Conf. Comput. Vis.})

@String(TOG= {ACM Trans. Graph.})

@String(ICLR = {Int. Conf. Learn. Represent.})

@String(VR   = {Vis. Res.})

@String(CVPR  = {CVPR})

@String(ICCV  = {ICCV})

@String(TOG   = {ACM TOG})

@String(ICLR  = {ICLR})


%% file: template.bib
@inproceedings{wang2025decoupledgaussian,
  title={Decoupledgaussian: Object-scene decoupling for physics-based interaction},
  author={Wang, Miaowei and Zhang, Yibo and Xu, Weiwei and Ma, Rui and Zou, Changqing and Morris, Daniel},
  booktitle={Proceedings of the Computer Vision and Pattern Recognition Conference},
  pages={11361--11372},
  year={2025}
}

@article{qiu2024feature,
  title={Feature splatting: Language-driven physics-based scene synthesis and editing},
  author={Qiu, Ri-Zhao and Yang, Ge and Zeng, Weijia and Wang, Xiaolong},
  journal={arXiv preprint arXiv:2404.01223},
  year={2024}
}

@inproceedings{silva_contrastive_2024,
  title={Contrastive gaussian clustering for weakly supervised 3d scene segmentation},
  author={Castillo, Myrna and Dahaghin, Mahtab and Toso, Matteo and Del Bue, Alessio},
  booktitle={International Conference on Pattern Recognition},
  pages={114--130},
  year={2024},
  organization={Springer}
}

@inproceedings{marrie2025ludvig,
  title={Ludvig: Learning-free uplifting of 2d visual features to gaussian splatting scenes},
  author={Marrie, Juliette and M{\'e}n{\'e}gaux, Romain and Arbel, Michael and Larlus, Diane and Mairal, Julien},
  booktitle={2025 IEEE/CVF International Conference on Computer Vision (ICCV)},
  pages={7440--7450},
  year={2025},
  organization={IEEE}
}

@inproceedings{chacko2025lifting,
  title={Lifting by gaussians: A simple, fast and flexible method for 3d instance segmentation},
  author={Chacko, Rohan and H{\"a}ni, Nicolai and Khaliullin, Eldar and Sun, Lin and Lee, Douglas},
  booktitle={2025 IEEE/CVF Winter Conference on Applications of Computer Vision (WACV)},
  pages={3497--3507},
  year={2025},
  organization={IEEE}
}

@article{schieber2026liftnav,
  title={LiftNav: Path Planning via Semantic Lifting in TSDF-Guided Gaussian Splatting},
  author={Schieber, Hannah and Frischmann, Dominik and Schaack, Victor and Schoellig, Angela P and Roth, Daniel},
  journal={arXiv preprint arXiv:2605.31376},
  year={2026}
}

@inproceedings{schonberger2016structure,
  title={Structure-from-motion revisited},
  author={Schonberger, Johannes L and Frahm, Jan-Michael},
  booktitle={Proceedings of the IEEE conference on computer vision and pattern recognition},
  pages={4104--4113},
  year={2016}
}

@article{meuleman2025fly,
  title={On-the-fly reconstruction for large-scale novel view synthesis from unposed images},
  author={Meuleman, Andreas and Shah, Ishaan and Lanvin, Alexandre and Kerbl, Bernhard and Drettakis, George},
  journal={ACM Transactions on Graphics (TOG)},
  volume={44},
  number={4},
  pages={1--14},
  year={2025},
  publisher={ACM New York, NY, USA}
}

@article{tang2026scar,
  title={SCAR-GS: Scene-level Contrastive Articulated Reconstruction via Actionable Semantic 3D Gaussian Splatting},
  author={Tang, Jiahao and Gu, Qiuyi and Yu, Jincheng and Su, Chen and Yi, Tinghao and Qiao, Fei and Wang, Yu},
  journal={IEEE Robotics and Automation Letters},
  year={2026},
  publisher={IEEE}
}

@inproceedings{zhou2024feature,
  title={Feature 3dgs: Supercharging 3d gaussian splatting to enable distilled feature fields},
  author={Zhou, Shijie and Chang, Haoran and Jiang, Sicheng and Fan, Zhiwen and Zhu, Zehao and Xu, Dejia and Chari, Pradyumna and You, Suya and Wang, Zhangyang and Kadambi, Achuta},
  booktitle={2024 IEEE/CVF Conference on Computer Vision and Pattern Recognition (CVPR)},
  pages={21676--21685},
  year={2024},
  organization={IEEE}
}

@article{guo2026semantic,
  title={Semantic gaussians: Open-vocabulary scene understanding with 3d gaussian splatting},
  author={Guo, Jun and Ma, Xiaojian and Fan, Yue and Liu, Huaping and Li, Qing},
  journal={IEEE Transactions on Circuits and Systems for Video Technology},
  year={2026},
  publisher={IEEE}
}

@inproceedings{ravi_sam_2024,
	title = {Sam 2: Segment anything in images and videos},
  author={Ravi, Nikhila and Gabeur, Valentin and Hu, Yuan-Ting and Hu, Ronghang and Ryali, Chaitanya and Ma, Tengyu and Khedr, Haitham and R{\"a}dle, Roman and Rolland, Chloe and Gustafson, Laura and others},
  booktitle={International Conference on Learning Representations (ICLR)},
  volume={2025},
  pages={28085--28128},
  year={2025}
}

@inproceedings{zhang2025cob,
  title={Cob-gs: Clear object boundaries in 3dgs segmentation based on boundary-adaptive gaussian splitting},
  author={Zhang, Jiaxin and Jiang, Junjun and Chen, Youyu and Jiang, Kui and Liu, Xianming},
  booktitle={2025 IEEE/CVF Conference on Computer Vision and Pattern Recognition (CVPR)},
  pages={19335--19344},
  year={2025},
  organization={IEEE}
}

@ARTICLE{schieber_splatxtract_2026,
  author={Schieber, Hannah and Kleinbeck, Constantin and Schoellig, Angela P. and Leutenegger, Stefan and Roth, Daniel},
  journal={IEEE Robotics and Automation Letters}, 
  title={SplatXtRact: Tractable Gaussian Splatting via Open World Region-of-Interest Extraction and Refinement}, 
  year={2026},
  volume={11},
  number={7},
  pages={8664-8671},
  doi={10.1109/LRA.2026.3699257}}

@inproceedings{qin_langsplat_2024,
	title = {{LangSplat}: 3D Language Gaussian Splatting},
	booktitle = {Proceedings of the {IEEE}/{CVF} Conference on Computer Vision and Pattern Recognition ({CVPR}) 2024},
	author = {Qin, Minghan and Li, Wanhua and Zhou, Jiawei and Wang, Haoqian and Pfister, Hanspeter},
	year = {2024},
}

@inproceedings{hasselman_arephotography_2023,
	title = {{ARephotography}: Revisiting Historical Photographs using Augmented Reality},
	pages = {1--7},
	booktitle = {Extended Abstracts of the 2023 {CHI} Conference on Human Factors in Computing Systems},
	author = {Hasselman, Tommy and Lo, Wei Hong and Langlotz, Tobias and Zollmann, Stefanie},
	year ={2023},
}

@inproceedings{yu_magnoramas_2021,
	title = {Magnoramas: Magnifying dioramas for precise annotations in asymmetric 3d teleconsultation},
	pages = {392--401},
	booktitle = {2021 {IEEE} Virtual Reality and 3D User Interfaces ({VR})},
	author = {Yu, Kevin and Winkler, Alexander and Pankratz, Frieder and Lazarovici, Marc and Wilhelm, Dirk and Eck, Ulrich and Roth, Daniel and Navab, Nassir},
	year = {2021},
}

@misc{kirillov_segment_2023,
	title = {Segment Anything},
  author={Kirillov, Alexander and Mintun, Eric and Ravi, Nikhila and Mao, Hanzi and Rolland, Chloe and Gustafson, Laura and Xiao, Tete and Whitehead, Spencer and Berg, Alexander C and Lo, Wan-Yen and others},
  booktitle={Proceedings of the IEEE/CVF international conference on computer vision},
  pages={4015--4026},
  year={2023}
}

@inproceedings{jiang2024vr,
  title={Vr-gs: A physical dynamics-aware interactive gaussian splatting system in virtual reality},
  author={Jiang, Ying and Yu, Chang and Xie, Tianyi and Li, Xuan and Feng, Yutao and Wang, Huamin and Li, Minchen and Lau, Henry and Gao, Feng and Yang, Yin and others},
  booktitle={ACM SIGGRAPH 2024 conference papers},
  pages={1--1},
  year={2024}
}

@article{chen2025splat,
  title={Splat-nav: Safe real-time robot navigation in gaussian splatting maps},
  author={Chen, Timothy and Shorinwa, Ola and Bruno, Joseph and Swann, Aiden and Yu, Javier and Zeng, Weijia and Nagami, Keiko and Dames, Philip and Schwager, Mac},
  journal={IEEE Transactions on Robotics},
  year={2025},
}

@article{chen2025grad,
  title={Grad-nav++: Vision-language model enabled visual drone navigation with gaussian radiance fields and differentiable dynamics},
  author={Chen, Qianzhong and Gao, Naixiang and Huang, Suning and Low, JunEn and Chen, Timothy and Sun, Jiankai and Schwager, Mac},
  journal={IEEE Robotics and Automation Letters},
  volume={11},
  number={2},
  pages={1418--1425},
  year={2025},
}

@article{fang2024mini,
  title={Mini-splatting2: Building 360 scenes within minutes via aggressive gaussian densification},
  author={Fang, Guangchi and Wang, Bing},
  journal={arXiv preprint arXiv:2411.12788},
  year={2024}
}

@article{jung_scrream_2025,
	title = {{SCRREAM}: {SCan}, Register, {REnder} And Map: A Framework for Annotating Accurate and Dense 3D Indoor Scenes with a Benchmark},
	volume = {37},
	pages = {44164--44176},
	journaltitle = {Advances in Neural Information Processing Systems},
	journal = {Advances in Neural Information Processing Systems},
	author = {Jung, HyunJun and Li, Weihang and Wu, Shun-Cheng and Bittner, William and Brasch, Nikolas and Song, Jifei and Pérez-Pellitero, Eduardo and Zhang, Zhensong and Moreau, Arthur and Navab, Nassir and {others}},
	year ={2025},
}

@inproceedings{yu_language-embedded_2024,
	title = {Language-embedded gaussian splats (legs): Incrementally building room-scale representations with a mobile robot},
	pages = {13326--13332},
	booktitle = {2024 {IEEE}/{RSJ} International Conference on Intelligent Robots and Systems ({IROS})},
	author = {Yu, Justin and Hari, Kush and Srinivas, Kishore and El-Refai, Karim and Rashid, Adam and Kim, Chung Min and Kerr, Justin and Cheng, Richard and Irshad, Muhammad Zubair and Balakrishna, Ashwin and {others}},
	year ={2024},
}

@misc{ji_graspsplats_2024,
	title = {Graspsplats: Efficient manipulation with 3d feature splatting},
	journaltitle = {{arXiv} preprint {arXiv}:2409.02084},
	author = {Ji, Mazeyu and Qiu, Ri-Zhao and Zou, Xueyan and Wang, Xiaolong},
	year = {2024},
}

@inproceedings{li_object-aware_2024,
	title = {Object-aware gaussian splatting for robotic manipulation},
	booktitle = {{ICRA} 2024 Workshop on 3D Visual Representations for Robot Manipulation},
	author = {Li, Yulong and Pathak, Deepak},
	year = {2024},
}

@inproceedings{huang_3d_2025,
	title = {3d gaussian inpainting with depth-guided cross-view consistency},
	pages = {26704--26713},
	booktitle = {Proceedings of the Computer Vision and Pattern Recognition Conference},
	author = {Huang, Sheng-Yu and Chou, Zi-Ting and Wang, Yu-Chiang Frank},
	year = {2025},
}

@inproceedings{weder_removing_2023,
	title = {Removing Objects From Neural Radiance Fields},
	pages = {16528--16538},
	booktitle = {Proceedings of the {IEEE}/{CVF} Conference on Computer Vision and Pattern Recognition ({CVPR})},
	author = {Weder, Silvan and Garcia-Hernando, Guillermo and Monszpart, Aron and Pollefeys, Marc and Brostow, Gabriel J. and Firman, Michael and Vicente, Sara},
	year = {2023-06},
}

@misc{yin_or-nerf_2023,
	title = {Or-nerf: Object removing from 3d scenes guided by multiview segmentation with neural radiance fields},
	journaltitle = {{arXiv} preprint {arXiv}:2305.10503},
	author = {Yin, Youtan and Fu, Zhoujie and Yang, Fan and Lin, Guosheng},
	year = {2023},
}

@article{hu_sagd_2024,
	title = {{SAGD}: Boundary-enhanced segment anything in 3D Gaussian via Gaussian decomposition},
	journal = {{arXiv} preprint {arXiv}:2401.17857},
	author = {Hu, Xu and Wang, Yuxi and Fan, Lue and Fan, Junsong and Peng, Junran and Lei, Zhen and Li, Qing and Zhang, Zhaoxiang},
	year = {2024},
}

@article{li_activesplat_2025,
	title = {Activesplat: High-fidelity scene reconstruction through active gaussian splatting},
	journaltitle = {{IEEE} Robotics and Automation Letters},
	journal = {{IEEE} Robotics and Automation Letters},
	author = {Li, Yuetao and Kuang, Zijia and Li, Ting and Hao, Qun and Yan, Zike and Zhou, Guyue and Zhang, Shaohui},
	year = {2025},
}

@inproceedings{tang_dronesplat_2025,
	title = {{DroneSplat}: 3D Gaussian Splatting for Robust 3D Reconstruction from In-the-Wild Drone Imagery},
	pages = {833--843},
	booktitle = {Proceedings of the Computer Vision and Pattern Recognition Conference ({CVPR})},
	author = {Tang, Jiadong and Gao, Yu and Yang, Dianyi and Yan, Liqi and Yue, Yufeng and Yang, Yi},
	year = {2025},
}

@misc{jiang_fisherrf_2023,
	title = {Fisherrf: Active view selection and uncertainty quantification for radiance fields using fisher information},
	journaltitle = {{arXiv} preprint {arXiv}:2311.17874},
	author = {Jiang, Wen and Lei, Boshu and Daniilidis, Kostas},
	year = {2023},
}

@inproceedings{irshad_neo_2023,
	title = {Neo 360: Neural fields for sparse view synthesis of outdoor scenes},
	pages = {9187--9198},
	booktitle = {Proceedings of the {IEEE}/{CVF} International Conference on Computer Vision},
	author = {Irshad, Muhammad Zubair and Zakharov, Sergey and Liu, Katherine and Guizilini, Vitor and Kollar, Thomas and Gaidon, Adrien and Kira, Zsolt and Ambrus, Rares},
	year = {2023},
}

@misc{qu_drag_2025,
	title = {Drag Your Gaussian: Effective Drag-Based Editing with Score Distillation for 3D Gaussian Splatting},
	journaltitle = {{arXiv} preprint {arXiv}:2501.18672},
	author = {Qu, Yansong and Chen, Dian and Li, Xinyang and Li, Xiaofan and Zhang, Shengchuan and Cao, Liujuan and Ji, Rongrong},
	year = {2025},
}

@misc{zhang_3ditscene_2024,
	title = {3DitScene: Editing Any Scene via Language-guided Disentangled Gaussian Splatting},
	journaltitle = {{arXiv} preprint {arXiv}:2405.18424},
	author = {Zhang, Qihang and Xu, Yinghao and Wang, Chaoyang and Lee, Hsin-Ying and Wetzstein, Gordon and Zhou, Bolei and Yang, Ceyuan},
	year = {2024},
}

@misc{lyu_gaga_2024,
	title = {Gaga: Group Any Gaussians via 3D-aware Memory Bank},
	author = {Lyu, Weijie and Li, Xueting and Kundu, Abhijit and Tsai, Yi-Hsuan and Yang, Ming-Hsuan},
	year = {2024},
	note = {\_eprint: 2404.07977},
}

@inproceedings{cheng_tracking_2023,
	title = {Tracking anything with decoupled video segmentation},
	pages = {1316--1326},
	booktitle = {Proceedings of the {IEEE}/{CVF} International Conference on Computer Vision},
	author = {Cheng, Ho Kei and Oh, Seoung Wug and Price, Brian and Schwing, Alexander and Lee, Joon-Young},
	year = {2023},
}

@inproceedings{ye_gaussian_2024,
	title = {Gaussian grouping: Segment and edit anything in 3d scenes},
	pages = {162--179},
	booktitle = {European Conference on Computer Vision},
	author = {Ye, Mingqiao and Danelljan, Martin and Yu, Fisher and Ke, Lei},
	year = {2024},
}

@inproceedings{schieber_semantics_controlled_2024,
	title = {Semantics-controlled gaussian splatting for outdoor scene reconstruction and rendering in virtual reality},
	pages = {318--328},
	booktitle = {2025 {IEEE} Conference Virtual Reality and 3D User Interfaces ({VR})},
	author = {Schieber, Hannah and Young, Jacob and Langlotz, Tobias and Zollmann, Stefanie and Roth, Daniel},
	year = {2025},
}

@article{li_cross-view_2025,
	title = {Cross-view geolocalization and disaster mapping with street-view and {VHR} satellite imagery: A case study of Hurricane {IAN}},
	volume = {220},
	pages = {841--854},
	journaltitle = {{ISPRS} Journal of Photogrammetry and Remote Sensing},
	journal = {{ISPRS} Journal of Photogrammetry and Remote Sensing},
	author = {Li, Hao and Deuser, Fabian and Yin, Wenping and Luo, Xuanshu and Walther, Paul and Mai, Gengchen and Huang, Wei and Werner, Martin},
	year = {2025},
}

@article{kerbl_3d_2023,
    title = {3d gaussian splatting for real-time radiance field rendering},
    volume = {42},
    number = {4},
    journal = {ACM Transactions on Graphics},
    author = {Kerbl, Bernhard and Kopanas, Georgios and Leimkühler, Thomas and Drettakis, George},
    year = {2023},
    pages = {1--14},
}

@inproceedings{kerr_lerf_2023,
    title = {Lerf: {Language} embedded radiance fields},
    booktitle = {Proceedings of the {IEEE}/{CVF} {International} {Conference} on {Computer} {Vision}},
    author = {Kerr, Justin and Kim, Chung Min and Goldberg, Ken and Kanazawa, Angjoo and Tancik, Matthew},
    year = {2023},
    pages = {19729--19739},
}

@article{kleinbeck_multi-layer_2024,
	title = {Multi-layer gaussian splatting for immersive anatomy visualization},
	journaltitle = {{IEEE} Transactions on Visualization and Computer Graphics},
	journal = {{IEEE} Transactions on Visualization and Computer Graphics},
	author = {Kleinbeck, Constantin and Schieber, Hannah and Engel, Klaus and Gutjahr, Ralf and Roth, Daniel},
	year = {2025},
}

@article{knapitsch_tanks_2017,
    title = {Tanks and temples: {Benchmarking} large-scale scene reconstruction},
    volume = {36},
    number = {4},
    journal = {ACM Transactions on Graphics (ToG)},
    author = {Knapitsch, Arno and Park, Jaesik and Zhou, Qian-Yi and Koltun, Vladlen},
    year = {2017},
    pages = {1--13},
}
